\documentclass[a4paper]{article}

\usepackage[authoryear,comma,longnamesfirst,sectionbib]{natbib}

\newcommand*{\ARXIV}{}%  % when uncommented it uses the *.bbl file 

\usepackage{comment}
\newcommand{\github}{\href{https://github.com/LeszekSzczecinski/LOP}{our GitHub repository}}

\newcommand{\CFilesBib}{Common.Files.Bib}

\usepackage{amsmath,amssymb,amscd,latexsym,dsfont,wasysym,bbm}
\usepackage{float}
\usepackage{color}
\usepackage{graphics}
\usepackage{graphicx}
\usepackage{subfigure}
\usepackage{multirow,multicol} 
\usepackage{verbatim}
\usepackage{psfrag}
\usepackage{comment}
\usepackage{makeidx}
\usepackage[]{algorithm}
\usepackage{algorithmicx}
\usepackage{algpseudocode}
\usepackage{cases}
 \usepackage{setspace}
\usepackage{pgfplots} %%%% added by LS for the errata 7.sept.2015
\newcommand{\tr}[1]{\textrm{#1}}
\newcommand{\mr}[1]{\mathrm{#1}}
\newcommand{\tnr}[1]{{\textnormal{#1}}}

\newcommand{\mc}[1]{\mathcal{#1}}
\newcommand{\mf}[1]{\mathsf{#1}}
 
\newcommand{\ms}[1]{\mathds{#1}}

\newcommand{\ov}[1]{\overline{#1}}

\newcommand{\un}[1]{\underline{#1}}
\newcommand{\ba}{\boldsymbol{a}}

\newcommand{\bone}{\boldsymbol{1}}
\newcommand{\brho}{\boldsymbol{\rho}}

\newcommand{\figref}[1]{Fig.~\ref{#1}}

\newcommand{\secref}[1]{Sec.~\ref{#1}}

\newcommand{\exref}[1]{Example~\ref{#1}}

\newcommand{\propref}[1]{Proposition~\ref{#1}}
\newcommand{\tabref}[1]{Table~\ref{#1}}

\newcommand{\ie}{i.e.,~} 		%note the comma and the space (~)
\newcommand{\eg}{e.g.,~}	%note the comma and the space (~)
\newcommand{\cf}{cf.~}		%note the space (~)

\renewcommand{\emptyset}{\varnothing} % Empty set \varnothing is nicer than \emptyset :)
\newcommand{\argmax}{\mathop{\mr{argmax}}}
\newcommand{\argmin}{\mathop{\mr{argmin}}}

\newcommand{\set}[1]{\{#1\}}
\newcommand{\SET}[1]{\left\{#1\right\}}

\newcommand{\cd}{\cdot}
\newcommand{\ld}{\ldots}

\newcommand{\nchoosek}[2]{{{#1} \choose {#2}}}
\newcommand{\PR}[1]{\Pr\SET{#1}}       	% Probability
\newcommand{\pdf}{f}            			% PDF. I vote for having the PDFs and the PMFs as italic because the pmf as \tr{P}_\bX(\bX) looks weird and also because I like to see the pmfs and the pdfs as functions, which we denote using italic letters (like the Q-function, f(t), s(t), h(t), etc.)

\newcommand{\IND}[1]{\ms{I}\big[{#1}\big]}   	% Indicator function
\newcommand{\Ex}{\ms{E}}     			% Expectation (AA).
\newcommand{\T}{^{\mf{T}}}            		% transpose
\newcommand{\dd}{\,\mr{d}}             		% LS: differentiation operator (\, added to make it nicer)
\newcommand{\mcA}{\mc{A}}

\newcommand{\mcF}{\mc{F}}

\newcommand{\mcI}{\mc{I}}

\newcommand{\mcR}{\mc{R}}

\newcommand{\mfP}{\mf{P}}

\newcommand{\Natural}{\mathbb{N}}		% N
\newcommand{\matW}{\tnr{\textbf{W}}}

\newcommand{\red}[1]{{\color{red}#1}}
\pgfplotsset{compat=1.12}
\usetikzlibrary{positioning,shadows,shapes,fit,arrows,backgrounds}
\tikzstyle{rect_my} = [draw, rectangle, minimum width=2cm, text width=1.8cm, fill=gray!15, 
  text centered,  minimum height=.9cm]
\tikzstyle{square_my} = [draw, rectangle, minimum width=1cm, text width=0.8cm, fill=gray!15, 
  text centered,  minimum height=.9cm]
\tikzstyle{square_my_graph} = [draw, rectangle, minimum width=1.2cm, text width=1cm, fill=gray!15, 
  text centered,  minimum height=1.2cm]
\tikzstyle{circle_my} = [draw, circle, minimum width=1cm, text width=0.8cm, fill=gray!15, 
  text centered,  minimum height=.9cm]
\tikzstyle{circle_my_graph} = [draw, circle, minimum width=1.1cm, text width=.8cm, fill=gray!15, 
  text centered]
\tikzstyle{cloud_my} = [draw, shape=cloud, minimum width=1cm, text width=0.8cm, fill=gray!15, 
  text centered,  minimum height=1cm]

\tikzstyle{point_my} = [draw=none, minimum width=0cm, text width=0cm, fill=none, 
  text centered,  minimum height=0cm]    
\tikzstyle{line_my} = [draw, -latex]    
\tikzstyle{box_my}=[draw, minimum size=2em, text width=4.5em, text centered]
\tikzstyle{bigbox_my}=[draw, inner sep=15pt]
\tikzstyle{arrow_my} = [thick,->,>=stealth]
\tikzstyle{noarrow_my} = [thick,-,=>stealth]

\newcommand{\sizf}{0.8}
\usepackage[acronym,nonumberlist]{glossaries} % make a separate list of acronyms
\usepackage{hyperref}
\usepackage{empheq}
\usepackage{verbatim}

\newacronym[\glsshortpluralkey=PDFs,\glslongpluralkey=probability density functions]{pdf}{PDF}{probability density function}
\newacronym[\glsshortpluralkey=CDFs,\glslongpluralkey=cumulative density functions]{cdf}{CDF}{cumulative density function}
\newacronym[\glsshortpluralkey=CCDFs,\glslongpluralkey=complementary cumulative density functions]{ccdf}{CDF}{complementary cumulative density function}
\newacronym[\glsshortpluralkey=PMFs,\glslongpluralkey=probability mass functions]{pmf}{PMF}{probability mass function}
\newacronym[]{lhs}{l.h.s.}{left-hand side}
\newacronym[]{rhs}{r.h.s.}{right-hand side} 

\newacronym[]{bicm}{BICM}{bit-interleaved coded modulation}
\newacronym[]{bicmid}{BICM-ID}{BICM with iterative demapping}
\newacronym[]{cm}{CM}{coded modulation}
\newacronym[]{tcm}{TCM}{trellis-coded modulation}
\newacronym[]{mlc}{MLC}{multi-level coding}
\newacronym[]{pam}{PAM}{pulse amplitude modulation}
\newacronym[]{bpsk}{BPSK}{binary phase shift keying}
\newacronym[]{qam}{QAM}{quadrature amplitude modulation}
\newacronym[]{16qam}{16-QAM}{16-points quadrature amplitude modulation}
\newacronym[]{psk}{PSK}{phase shift keying}
\newacronym[\glsshortpluralkey=LLRs,\glslongpluralkey=logarithmic likelihood ratios]{llr}{LLR}{logarithmic likelihood ratio}
\newacronym[]{oc}{OC}{operating characteristic}
\newacronym[]{dmp}{DMP}{discretized message passing}
\newacronym[]{mp}{MP}{message passing}
\newacronym[]{ep}{EP}{expectation propagation}
\newacronym[\glsshortpluralkey=MIs,\glslongpluralkey=mutual informations]{mi}{MI}{mutual information}
\newacronym[\glsshortpluralkey=GMIs,\glslongpluralkey=generalized mutual informations]{gmi}{GMI}{generalized mutual information}
\newacronym[]{eesm}{EESM}{exponential effective-SNR-mapping}
\newacronym[]{bicm-gmi}{BICM-GMI}{BICM generalized mutual information}
\newacronym[]{awgn}{AWGN}{additive white Gaussian noise}
\newacronym[]{bsc}{BSC}{binary symetric channel}
\newacronym[]{amc}{AMC}{adaptive modulation and coding}
\newacronym[]{csi}{CSI}{channel state information}
\newacronym[]{cqi}{CQI}{channel quality indicator}
\newacronym[]{kl}{KL}{Kullback-Leibler}
\newacronym[]{cmm}{CMM}{circular moment matching}
\newacronym[]{ga}{GA}{Gaussian approximation}

\newacronym[]{sp}{SP}{set-partitioning}
\newacronym[]{gsm}{GSM}{global system for mobile communications}
\newacronym[]{edge}{EDGE}{enhanced data rates for GSM evolution}
\newacronym[]{3gpp}{3GPP}{3rd generation partnership project}
\newacronym[]{umts}{UMTS}{Universal Mobile Telecommunication System}
\newacronym[]{lte}{LTE}{Long Term Evolution}
\newacronym[]{dvb}{DVB}{digital video broadcasting}
\newacronym[]{fdd}{FDD}{Frequency Division Duplexing}

\newacronym[\glsshortpluralkey=CCs,\glslongpluralkey=convolutional codes]{cc}{CC}{convolutional code}
\newacronym[\glsshortpluralkey=PCCCs,\glslongpluralkey=parallel concatenated convolutional codes]{pccc}{PCCC}{parallel concatenated convolutional code}
\newacronym[\glsshortpluralkey=TCs,\glslongpluralkey=turbo codes]{tc}{TC}{turbo code}
\newacronym{ldpc}{LDPC}{low-density parity-check}
\newacronym[]{ofdm}{OFDM}{orthogonal frequency-division multiplexing}
\newacronym[]{bep}{BEP}{bit-error probability}
\newacronym[]{wep}{WEP}{word-error probability}
\newacronym[]{sep}{SEP}{symbol-error probability}
\newacronym[]{pep}{PEP}{pairwise-error probability}
\newacronym[]{ttcm}{TTCM}{turbo-trellis coded modulation}
\newacronym[]{uep}{UEP}{unequal error protection}
\newacronym[\glsshortpluralkey=CENCs,\glslongpluralkey=convolutional encoders]{cenc}{CENC}{convolutional encoder}
\newacronym[]{mimo}{MIMO}{multiple-input multiple-output}
\newacronym[\glsshortpluralkey=SNRs,\glslongpluralkey=signal-to-noise ratios]{snr}{SNR}{signal-to-noise ratio}
\newacronym[\glsshortpluralkey=SINRs,\glslongpluralkey=the signal-to-interference-plus-noise ratios]{sinr}{SINR}{the signal-to-interference-plus-noise ratio}
\newacronym[]{msb}{MSB}{most-significative bit}
\newacronym[]{bcjr}{BCJR}{Bahl--Cocke--Jelinek--Raviv}
\newacronym[]{cbc}{CBC}{Colavolpe--Barbieri--Caire}
\newacronym[]{skr}{SKR}{Shayovitz--Kreimer--Raphaeli}
\newacronym[\glsshortpluralkey=SEDs,\glslongpluralkey=squared Euclidean distances]{sed}{SED}{squared Euclidean distance}
\newacronym[\glsshortpluralkey=EDs,\glslongpluralkey=Euclidean distances]{ed}{ED}{Euclidean distance}
\newacronym[\glsshortpluralkey=MEDs,\glslongpluralkey=minimum Euclidean distances]{med}{MED}{minimum Euclidean distance}
\newacronym[]{core}{CoRe}{constellation rearrangement}
\newacronym[]{pdl}{PDL}{parallel decoding of the individual levels}
\newacronym[\glsshortpluralkey=GCs,\glslongpluralkey=Gray codes]{gc}{GC}{Gray code}
\newacronym[]{brgc}{BRGC}{binary-reflected Gray code}
\newacronym[]{nbc}{NBC}{natural binary code}
\newacronym[]{fbc}{FBC}{folded-binary code}
\newacronym[]{bsgc}{BSGC}{binary semi-Gray code}
\newacronym[]{msp}{MSP}{modified set-partitioning}
\newacronym[]{ssp}{SSP}{semi set-partitioning}
\newacronym[]{fhd}{FHD}{free Hamming distance}
\newacronym[]{mfhd}{MFHD}{maximum free Hamming distance}
\newacronym[]{ods}{ODS}{optimal distance spectrum}
\newacronym[]{iud}{i.u.d.}{independent and uniformly distributed}
\newacronym[]{ud}{u.d.}{uniformly distributed}
\newacronym[]{iid}{i.i.d.}{independent, identically distributed}
\newacronym[]{ami}{AMI}{accumulated mutual information}
\newacronym[]{bico}{BICO}{binary-input continuous-output}
\newacronym[]{gh}{GH}{Gauss--Hermite}
\newacronym[]{gum}{GUM}{Gaussian--uniform mixture}
\newacronym[\glsshortpluralkey=BSs,\glslongpluralkey=base-stations]{bs}{BS}{base-station}
\newacronym[\glsshortpluralkey=MSs,\glslongpluralkey=mobile-stations]{ms}{MS}{mobile-stations}

\newacronym[]{phy}{PHY}{physical layer} 
\newacronym[]{rlc}{RLC}{Radio-Link control} 
\newacronym[]{ran}{RAN}{Radio Access Network} 
\newacronym[]{llc}{LLC}{logical link control} 
\newacronym[]{tcp}{TCP}{transmission control protocol} 
\newacronym[]{mac}{MAC}{media access control} 
\newacronym[]{fft}{FFT}{fast Fourier transform} 
\newacronym[]{ft}{FT}{Fourrier transform}
\newacronym[]{cf}{CF}{characteristic function} 
\newacronym[]{mgf}{MGF}{moment generating function} 
\newacronym[]{ee}{EE}{energy efficiency} 
\newacronym[]{eb}{EB}{energy per bit}
\newacronym[]{kkt}{KKT}{Karush--Kuhn--Tucker} 
\newacronym[]{mcs}{MCS}{modulation/coding scheme} 
\newacronym[]{fec}{FEC}{forward error correction}
\newacronym[]{arq}{ARQ}{automatic repeat request}
\newacronym[]{harq}{HARQ}{hybrid ARQ}
\newacronym[]{tarq}{TARQ}{truncated HARQ}
\newacronym[]{ir}{IR}{incremental redundancy}
\newacronym[]{rpr}{RR}{repetition redundancy}
\newacronym[]{rrharq}{RR-HARQ}{repetition redundancy HARQ}
\newacronym[]{irharq}{IR-HARQ}{incremental redundancy HARQ}
\newacronym[]{ack}{ACK}{positive acknowledgment}
\newacronym[]{nack}{NACK}{negative acknowledgment}
\newacronym[]{hol}{HoL}{head of the line}
\newacronym[]{crc}{CRC}{cyclic redundancy check}
\newacronym[]{dp}{DP}{dynamic programming}
\newacronym[]{gp}{GP}{geometric programming}
\newacronym[]{per}{PER}{packet error rate}
\newacronym[]{ber}{BER}{bit error rate}
\newacronym[]{op}{OP}{outage probability}
\newacronym[]{spa}{SPA}{saddle-point approximation}
\newacronym[]{mrc}{MRC}{maximum ratio combining}
\newacronym[]{mdp}{MDP}{Markov decision process}
\newacronym[]{lp}{LP}{linear programming}
\newacronym[]{pomdp}{POMDP}{partially observable Markov decision process}
\newacronym[]{psimdp}{PSI-MDP}{partial state information Markov decision process}
\newacronym[]{scpp}{SCPP}{stochastic shortest path problem}

\newacronym[]{forw}{frwd}{forward}
\newacronym[]{feed}{fdbk}{feedback}

\newacronym[]{mm}{MM-HARQ}{multi-message HARQ}
\newacronym[]{xp}{XP-HARQ}{cross-packet HARQ}
\newacronym[]{ts}{TS}{time-sharing}
\newacronym[]{sc}{SC}{superposition coding}
\newacronym[]{sbrq}{SBRQ}{systematic backward retransmission}
\newacronym[]{brq}{BRQ}{backward retransmission}
\newacronym[]{lharq}{L-HARQ}{layer-coded HARQ}
\newacronym[]{anlharq}{AoN-HARQ}{all-or-none L-HARQ}
\newacronym[]{vlharq}{VL-HARQ}{variable-length HARQ}

\newacronym[]{pp}{PPP}{point process}
\newacronym[]{ppp}{PPP}{Poisson point process}

\newacronym[]{fide}{FIDE}{F\'ed\'eration Internationale des \'Echecs}
\newacronym[]{fifa}{FIFA}{F\'ed\'eration Internationale de Football Association}
\newacronym[]{fivb}{FIVB}{F\'ed\'eration Internationale de Volleyball}
\newacronym[]{epl}{EPL}{English Premier League}
\newacronym[]{nhl}{NHL}{National Hockey League}
\newacronym[]{nfl}{NFL}{National Football League}
\newacronym[]{ipl}{IPL}{Indian Premier League}

\newacronym[]{sg}{SG}{stochastic gradient}
\newacronym[]{lms}{LMS}{least mean squares}
\newacronym[]{rls}{RLS}{recursive least squares}
\newacronym[]{vss}{VSS}{variable step-size}
\newacronym[]{hfa}{HFA}{home-field advantage}
\newacronym[]{ha}{HA}{home advantage}
\newacronym[]{mov}{MOV}{margin of victory}
\newacronym[]{ac}{AC}{Adjacent Categories}
\newacronym[]{cl}{CL}{Cumulative Link}
\newacronym[]{rps}{RPS}{Ranked Probability Score}
\newacronym[]{mse}{MSE}{Mean Squared Error}
\newacronym[]{mmse}{MMSE}{Minimum Mean Squared Error}
\newacronym[]{rmse}{RMSE}{Root Mean Squared Error}
\newacronym[]{map}{MAP}{maximum a posteriori}
\newacronym[]{ml}{ML}{maximum likelihood}
\newacronym[]{loo}{LOO}{leave-one-out}
\newacronym[]{alo}{ALO}{approximate leave-one-out}
\newacronym[]{logo}{LOGO}{leave-one-game-out}
\newacronym[]{alogo}{ALOGO}{approximate leave-one-game-out}
\newacronym[]{msd}{MSD}{mean-square deviation}
\newacronym[]{lop}{LOP}{linear ordering problem}

\newacronym[]{svd}{SVD}{singular values decomposition}

\newacronym[]{skf}{SKF}{Simplified Kalman Filter}
\newacronym[]{vskf}{vSKF}{\emph{vector-covariance} Simplified Kalman Filter}
\newacronym[]{sskf}{sSKF}{\emph{scalar-covariance} Simplified Kalman Filter}
\newacronym[]{fskf}{fSKF}{\emph{fixed-variance} Simplified Kalman Filter}
\newacronym[]{kf}{KF}{Kalman Filter}
\newacronym[]{gelo}{G-Elo}{Generalized Elo}

\newacronym[]{tpb}{TPB}{tensor-product-basis}

\usepackage{tikz,pgfplots}
\pgfplotsset{width=3.4in,height=2.3in,compat=1.12} %% dimension of the figure
\pgfplotsset{width=3.4in,height=2.3in} %% dimension of the figure
\usetikzlibrary{positioning,shadows,shapes,fit,arrows,backgrounds}

\newtheorem{proposition}{Proposition}

\newtheorem{conjecture}{Conjecture}
\newtheorem{example}{Example}

\providecommand{\keywords}[1]
{
  \small	
  \textbf{\textit{Keywords---}} #1
}

\usepackage{authblk}

\title{Rankability and Linear Ordering Problem:\\Probabilistic Insight and Algorithms}
\author[1]{Leszek Szczecinski\thanks{Corresponding author: leszek.szczecinski@inrs.ca}}
\affil[1]{Institut National de la Recherche  Scientifique, Montreal, QC, H5A 1K6, Canada.}
\author[2]{Harsh Sukheja\thanks{harshanilsukheja@gmail.com}}
\affil[2]{Indian Institute of Technology Kharagpur, Kharagpur - 721302, West Bengal, India.}

\begin{document}

\maketitle

\begin{abstract}
The \acrfull{lop}, which consists in ordering $M$ objects from their pairwise comparisons, is commonly applied in many areas of research. While efforts have been made to devise efficient \acrshort{lop} algorithms, verification of whether the data are \emph{rankable}, that is, if the \gls{lop} solutions have a meaningful interpretation, received much less attention. To address this problem, we adopt a probabilistic perspective where the results of pairwise comparisons are modeled as Bernoulli variables with a common parameter and we estimate the latter from the observed data. The brute-force approach to the required enumeration has a prohibitive complexity of $O(M!)$, so we reformulate the problem and introduce a concept of the Slater spectrum that generalizes the Slater index, and then devise an algorithm to find the spectrum with complexity $O(M^3 2^M)$ that is manageable for moderate values of $M$. Furthermore, with a minor modification of the algorithm, we are able to find \emph{all} solutions of the \acrshort{lop} with the complexity $O(M2^M)$. Numerical examples are shown on synthetic and real-world data, and the algorithms are publicly available.
\end{abstract}

% \begin{keywords}
% \end{keywords}

\section{Introduction}
In this work, we propose models and methods to analyze the rankability of the data which tells us if it is possible to find the order among objects by solving the \gls{lop}. %The latter finds an order  from the results of objects' pairwise comparisons. 
The \gls{lop} has found wide use in many different areas, \eg for the aggregation of the preferences of individual judgments or for ranking in sports where the teams (or players) are compared through the game result. Many more applications rely on the \gls{lop} formulation, see \cite[Sec.~1]{Ceberio15}, \cite[Sec.~1]{Marti12}, or \cite{Charon10}; so, while the wording can change from one application to another, the fundamental problem remains the same.

The effort in the area of the \gls{lop} was mainly focused on devising efficient numerical solutions as, even if the \gls{lop} can be solved exactly \citep{deCani69}, for large problems, approximations are unavoidable due to its NP-hardness; see \citep{Charon10}, \citep{Marti12}, \citep{Ceberio15} for an overview of the literature.

Recent works \eg \cite{Anderson19}, \cite{Anderson21}, \cite{Cameron21} addressed an important issue of data \emph{rankability}, that is, deciding if the data can be ordered. This is a critical step because, while objects can be ordered algorithmically, we need to know whether the ranking obtained can be meaningfully interpreted.

We continue this line of research, but, unlike \cite{Anderson19} or other works, we use a probabilistic approach, explicitly modeling the results of pairwise comparisons as Bernoulli variables. This follows the spirit of the early analysis in \cite{Slater61} or \cite{Remage66}. In particular, \cite{Slater61} proposed a hypothesis testing approach, which assumes that all Bernoulli variables have the same distribution with parameter $p$, and verifies if hypothesis $p=0.5$ (meaning that the data is purely random and not related to the order) can be rejected. However, this method is too complex to be practical in most cases. 

We also use the assumption that $p$ is common to all variables, but instead of testing the hypothesis, we estimate $p$ from the data. The estimation problem relies on finding the so-called Slater spectrum, which generalizes the well-known Slater index and tells us how ``far'' all possible orders are from the results we observe. We devise an algorithm that allows us to obtain the Slater spectrum efficiently. This is our main contribution. Incidentally, the algorithm we develop also allows us to find all the solutions of the \gls{lop}, which is itself an interesting contribution because we show that it can be done with the complexity $O(M2^M)$ which is known to be the bound on the complexity of finding only one solution \citep{Remage66}.

The remainder of this paper is organized as follows. In \secref{Sec:Model} we describe the problem, define the model, show the relationship between the \gls{lop} and its probabilistic counterpart, and briefly describe the approach suggested in \cite{Slater61}. The estimation problem is defined in \secref{Sec:Estimation} while \secref{Sec:Recursive.Calculation} focuses on the proposed algorithm and evaluates the related numerical complexity. Interestingly, as a ``byproduct" of the algorithm developed, we are able to recover all the solutions of the \gls{lop} and this comes with virtually no additional cost; the additional algorithmic steps required to do that are explained in \secref{Sec:All.Solutions}. The algorithms implemented in Python are available at \github. Numerical examples are shown in \secref{Sec:Numerical.Examples} where we use synthetic data, as well as the results of cricket games in the \acrfull{ipl} and the volleyball games of the Italian SuperLega. We conclude the work in \secref{Sec:Conclusions}.

%%%%%%%%%%%%%%%%%%%%%%%%%%%%%%%%%%%%%%%%%%%%%%%%%%%%%%%%%%%
\section{Model and problem definition}\label{Sec:Model}

Consider objects indexed with $M$ integers from the set $\mcI=\set{1,\ld, M}$ and assume that they can be ordered, \eg $2\succ 5 \succ 1 \succ\ld \succ M \succ 4$, where $m\succ n$ means that object $m$ is ordered \emph{before} object $n$ and the usual meaning is that ``$m$ is preferred to $n$" (in social science), or, in sports, that ``team $m$ is better than team $n$". The order will be denoted by $\brho=[\rho_1,\rho_2,\ld,\rho_M]\in\mfP_{\mcI}$, where $\rho_i\in\mcI, i=1,\ld,M$, $\rho_1 \succ \rho_2\succ\ld,\succ \rho_M$, and $\mfP_{\mcI}$ is the space of permutations of $\mcI$, $|\mfP_{\mcI}|=M!$. It essentially means that the object $\rho_l$ is ordered at position $l$, \eg in sport context, $\rho_1$ is the ``best'' team/player.  In the previous example, we have $\rho_1=2$, $\rho_2=5$, $\rho_3=1$, $\rho_{M-1}=M$, and $\rho_M=4$.

The order $\brho$, being unknown, can only be inferred from the observations and the inference result, $\hat\brho$ is called a \emph{ranking}.
We assume that we can observe the results of comparison between a pair of objects, where the simplest case occurs if the result of the comparison is binary, \ie either $m$ wins against $n$, or $n$ wins against $m$, that is, the ties are not allowed. In this case, the results can be stored in the \emph{observation} matrix $\matW$ with the entries $w_{m,n}\in\Natural$ indicating how many times $m$ won against $n$. Diagonal entries in the matrix are ignored or, equivalently, set to zero, $w_{m,m}=0$. The total number of comparisons is $T=\sum_{m,n} w_{m,n}$ and the number of comparisons between $m$ and $n$ is given by $K_{m,n}=K_{n,m}=w_{m,n}+w_{n,m}$.

The total number of distinct pairs is given by $\nchoosek{M}{2}$. In general, not all pairs need to be compared, and the comparison may be performed multiple times for the same pair. For example, in sports, some players face each other very rarely or even never, while others compete frequently.

If knowing $m\succ n$ implied that $w_{n,m}=0$, the pairwise comparison would be in perfect agreement with the underlying order of the objects. In practice, however, we may have both $w_{m,n}>0$ and $w_{n,m}>0$, which means that, in some comparisons (between $m$ and $n$) $m$ wins and, in others, $n$ does. This inconsistency of observations with the underlying order may be interpreted as a manifestation of randomness that may be modelled using Bernoulli variables:
\begin{align}
\label{Pr=p}
    \PR{m \tnr{ wins against } n| m\succ n} &= p\\
\label{Pr=q}
    \PR{m \tnr{ wins against } n| m\prec n} &= 1-p, 
\end{align}
where $p\in[0.5,1]$. At the extremes, $p=1$ means that observations are always in agreement with the underlying order, while $p=0.5$ means that the win of $m$ over $n$ is independent of the order: in this case, searching for the order is obviously useless, which is synonymous with the data being not rankable. 

We may estimate the ranking via \gls{ml} 
\begin{align}
\label{ML.definition}
    \hat\brho &= \argmax_{\brho\in\mfP_{\mcI}} J(\brho)\\
    % J(\brho) & = \log\PR{\matW | \brho}
\label{J.brho.definition}
    J(\brho) & = \PR{\matW | \brho}
\end{align}
where, in general, \eqref{ML.definition} may have many equivalent solutions $\hat\brho$ \citep{Anderson22}. 

For a known/given order $\brho$, the results (comparisons) are mutually independent; this conventional modeling assumption, which allows us to write \eqref{J.brho.definition} as follows:
\begin{align}
    J(\brho) 
    &=  \prod_{m\neq n}  \PR{w_{m,n}|\brho}\\
\label{J.brho}
             & = p^{T- S(\brho)} (1-p)^{S(\brho)},\\
\label{C.brho}
    S(\brho) &= T-C(\brho),\\
\label{Consistency.brho}
    C(\brho) &= \sum_{m=1}^M\sum_{n=m+1}^M w_{\rho_m,\rho_n}
\end{align}
where the notation $\PR{w_{m,n}|\brho}$ should be read as the probability of observing $w_{m,n}$ wins of $m$ over $n$ conditioned on the given order $\brho$

The \emph{inconsistency index} $S(\brho)$ denotes the number of observations that are in disagreement with the order $\brho$. For example, in sports, $S(\brho)$ is the number of games in which favorite teams lose against the underdogs (where, according to $\brho$, the favorite is ordered before the underdog). Similarly, the \emph{consistency index} $C(\brho)$ measures the number of observations in agreement with the order $\brho$. 

An alternative formulation of the \gls{lop}, known as \emph{matrix triangulation}, sees \eqref{ML.definition.final.C} as a search for a permutation of rows and columns of the matrix $\matW$, which maximizes $C(\brho)$ which is the sum of elements in the upper triangular part of the permuted matrix $\matW$; this can be seen directly in \eqref{Consistency.brho}.

As long as $p>0.5$, we may write \eqref{ML.definition} as
\begin{align}
\label{ML.definition.final.C}
    \hat\brho &= \argmin_{\brho\in\mfP_{\mcI}} S(\brho)=\argmax_{\brho\in\mfP_{\mcI}} C(\brho)
\end{align}
and, the inconsistency index of the optimal ranking $\hat\brho$
\begin{align}
\label{ML.definition.final.C.hat}
    \hat{S} & = S(\hat\brho),
\end{align}
is known as the Slater index \citep{Charon10}.

Therefore, for all $p>0.5$, there is a one-to-one mapping between problems \eqref{ML.definition.final.C} and \eqref{ML.definition}, which explains the appeal of the model \eqref{Pr=p}-\eqref{Pr=q}: we do not need to \emph{know} $p$ to find the ranking $\hat\brho$, and then we may focus on finding numerically efficient methods to solve \eqref{ML.definition.final.C}. 

It should be noted that the literature may formulate a more general version of \eqref{Pr=p}-\eqref{Pr=q}, by assuming that each comparison of $m$ vs. $n$ has its own parameter $p_{m,n}$ which replaces $p$ in \eqref{Pr=p} \citep{deCani69}, \citep{Flueck75}, \citep{Tiwisina19}. However, then, to find $\hat\brho$, we \emph{must} know or estimate $M(M-1)/2$ parameters $p_{m,n}$ \citep{deCani69} -- a problem compounded by the fact that additional transitivity conditions must be satisfied $m\succ k\succ n \implies p_{m,n}\ge p_{m,k} \wedge p_{m,n}\ge p_{k,n}$, \citep{Tiwisina19}.

The optimization is then quite difficult, \eg \citet{Tiwisina19} needed heuristic optimization tools. However, in addition to numerical problems, difficulty may appear due to the fact that the number of parameters to estimate may be relatively large compared to the number of observations $T$. For the sake of argument, assuming (very unrealistically) that we \emph{know} $\brho$, with $K_{m,n}=K$, there are $K$ observations per parameter. This affects the reliability of the estimate as, in practice, $K$ may be small. For example, using football data, as done in \cite{Tiwisina19}, we would have $K=2$ observations per each Bernoulli parameter $p_{m,n}$; then, the reliability of the estimation of $p_{m,n}$ may and should be questioned.

Taking into account the above, using a model \eqref{Pr=p}-\eqref{Pr=q} may be seen as a pragmatic middle ground, striking a balance between estimation complexity/reliability and the generality of a model. However, our goal is not to discuss the modeling assumption, but rather to point out the fact that many \gls{lop}s are of the form \eqref{ML.definition.final.C} and thus, implicitly, are related to the model \eqref{Pr=p}-\eqref{Pr=q}.

%%%%%%%%%%%%%%%
\subsection{Rankability}\label{Sec:Rankability}

The equivalence between \eqref{ML.definition.final.C} and \eqref{ML.definition} requires $p>0.5$. On the other hand, for $p=0.5$, \ie when the data is not rankable, we have $J(\brho)\equiv 2^{-T}$. Then, we can still find solutions of \eqref{ML.definition.final.C}, but they do not solve \eqref{ML.definition} in any meaningful way simply because $J(\brho)$ is independent of $S(\brho)$. Therefore, we conclude that the rankability of the data $\matW$ is captured by the parameter $p$, and our work focuses on this very issue; we will calculate the statistics of $p$, and more generally obtain its posterior distribution, $\pdf(p|\matW)$.

This approach to rankability follows the postulate of  \citet[Sec.~1]{Anderson19} which requires the rankability criterion to be agnostic with respect to the ranking algorithm. Indeed, to conclude if the data $\matW$ is rankable, we find $\pdf(p|\matW)$ and this does not presuppose how the ranking $\hat\brho$ should be found. In fact, as we shall see, the solution $\hat{\brho}$ found in \eqref{ML.definition.final.C} does not play a particular role, and we scan the whole permutation space $\mf{P}_{\mcI}$. Before explaining our approach, we can revisit some previous takes on rankability.

%%%%%%%%%%%%%%%%%%%
\subsubsection{Slater's method}\label{Sec:Slaters.index}

The rankability has been addressed in \cite{Slater61} through hypothesis testing where the hypothesis $H_0$ is that all $T$ observations are obtained with probability $p=0.5$, that is, are drawn uniformly from the space of $2^T$ elements, each represented by the observation matrix which we can treat as a random variable $\un\matW$. 

The rankings obtained from random observations $\un\matW$ are also random
\begin{align}
    \hat{\brho}_{\un\matW} &= \argmin_{\brho\in\mfP_{\mcI}} S(\brho; \un\matW)\\
    \hat{\un{S}}&=S(\hat\brho_{\un{\matW}};\un{\matW}),
\end{align}
where $S(\brho,\matW)$ is the same as \eqref{C.brho} and $\brho_{\matW}$ is the same as \eqref{ML.definition} with the dependence on the observation matrix explicitly indicated.

If, under hypothesis $H_0$, the probability of obtaining the Slater index $\hat{\un{S}}$ smaller than $\hat{S}$ is negligible, we reject $H_0$, \ie the data is considered rankable. More precisely, we calculate
\begin{align}
\label{testing.H0}
    p_{\tr{val}}=\mr{Pr}_{\un{\hat{S}}}\set{\un{\hat{S}}\le \hat{S}}
\end{align}
and compare it to a small threshold $\epsilon>0$, \eg it is common to use $\epsilon = 0.05$. Then
\begin{align}
    p_{\tr{val}}>\epsilon\implies \tr{reject $H_0$}.
\end{align}

We note that \cite{Slater61} suggested calculating \eqref{testing.H0} by brute force, \ie by (i) enumerating all possible outcomes, (ii) forming the corresponding observation matrices $\matW$, (iii) solving \eqref{ML.definition.final.C} for each of the matrices, and (iv) counting all solutions whose Slater index is smaller than $\hat{S}$.

The computational bottleneck lies in steps (i)-(iii): solving the \gls{lop} for all realizations of the matrices $\matW$, is too complex to be practical. For example, with the simplest scheduling $K_{m,n}=1$ (\ie one comparison per pair), we have $T=\nchoosek{M}{2}=M(M-1)/2$ and thus we have to solve the \gls{lop} for $2^T$ matrices $\matW$. Even if $M$ is moderate, \eg for $M=20$, when \eqref{ML.definition.final.C} is easily solved, solving it $2^T\approx 10^{57}$ times is an impossible challenge. 

Thus, the above explicit implementation is not applicable in practice. Alternatively, we may evaluate the probability in \eqref{testing.H0} via the Monte Carlo method, by mimicking steps (i)-(iv) shown above. Then, in steps (i)-(ii), instead of enumeration, we will randomly generate $N_{\tr{MC}}$ matrices $\matW$ with probability $p=0.5$. Of course, we still have to solve \eqref{ML.definition.final.C} $N_{\tr{MC}}$ times.

The main theoretical issue with Slater' method is that $\hat{S}$ \emph{is not} a sufficient statistic for $p$, which simply means that, using only $\hat{S}$, we lose information about $\pdf(p|\matW)$ -- the posterior distribution of $p$. Consequently, any inference about $p$, made using only $\hat{S}$, will be an approximation. This will become obvious in \ref{Sec:Spectrum} where we will show a sufficient statistic.

%Another issue with this method is that it does not provide the \emph{value} of $p$ that would characterize well the problem. The objective of our work is to address the latter problem and propose numerically efficient methods to solve it. 

%%%%%%%%%%%%%%%%%%%%%%%%%%%%%%%%%%%%%%%%%%%%%%%%%%%%%%%%
\subsubsection{Optimal solutions}\label{Sec:property.opt.solution}

Because Slater's method is difficult to apply in practice, the rankability problem has been addressed using the quantitative metrics related to the properties of the solution $\hat\brho$, \eg in \cite{Anderson19}, \cite{Anderson21}, \cite{Cameron21}, \cite{Anderson22}.

The simplest to understand is the \emph{degree of linearity} \citep[Sec.~3.1]{Cameron21}:
\begin{align}
\label{degree.linearity}
    \lambda = 1-\frac{\hat{S}}{T},
\end{align}
which expresses the proportion of observations in $\matW$ which are consistent with the optimal ranking $\hat\brho$.

Other metrics, \eg shown in \cite{Anderson19}, \cite{Anderson21}, \cite{Cameron21}, \cite{Anderson22} are more difficult to calculate and we do not detail them here. Nevertheless, from the perspective of our work, these metrics $r$ may be written as
\begin{align}
    r = r( \hat\mcR )
\end{align}
where $\hat{\mcR}=\set{\brho: S(\brho)=\hat{S}}$ is the set of all optimal solutions $\hat\brho$. In particular, some of these metrics $r(\hat{\mcR})$ take into account not only $\hat{S}$ but also all solutions $\brho\in\hat{\mcR}$. 

Although this should offer an improvement over the criterion in \eqref{degree.linearity}, and may provide an interpretation to the existence of multiple solutions \cite{Anderson22}, it shares the main problem of Slater's method by using insufficient statistics of $p$.

%%%%%%%%%%%%%%%%%%%%%%%%%%%%%%%%%%%%%%%%%%%%%%%%%%%%%%%%
\section{Probabilistic inference}\label{Sec:Estimation}

The approach we propose can be seen as an estimation of the Bernoulli distribution parameter, $p$, from the observation matrix $\matW$.

We quickly note that it is tempting to solve the joint \gls{ml} problem
\begin{align}
\label{joint.ML}
    \hat{p}, \hat\brho = \argmax_{p, \brho} J_{p}(\brho),
\end{align}
where $J_{p}(\brho)$ is the same as \eqref{J.brho} with the dependence on $p$ explicitly indicated.

But since $\hat\brho$ does not depend on $p$, we have $\hat{S}=S(\hat\brho)$ so
\begin{align}
\label{joint.ML.definition}
    \hat{p} 
        % &= \argmax_{p} J_{p}(\hat\brho)\\
        &= \argmax_{p} p^{T-\hat{S}} (1-p)^{\hat{S}}\\
    \label{joint.ML.alpha.only}
        &= 1-\frac{\hat{S}}{T}=\lambda;
\end{align}
thus, the degree of linearity, $\lambda$, shown in \eqref{degree.linearity} is the estimate of $\hat{p}$ in the joint \gls{ml} problems \eqref{joint.ML}.

As we said in \secref{Sec:Rankability}, $\hat{S}$ and $\lambda$ are not, in general, sufficient statistics for $p$.

The problem with the joint \gls{ml} estimation is that, focusing on the mode, it ignores the form of the underlying distribution. Here, we propose to find the latter
\begin{align}
\label{aposteriori.W.p.p}
    \pdf(p|\matW)
    &\propto
    \Pr(\matW|p)\pdf(p)\\
\label{likelihood.W.p}
    &\propto
    \Pr(\matW|p)\\
\label{Pr.W.rho.p.marginalization}
    &=\sum_{\brho \in \mfP_{\mcI}} \PR{\matW|p,\brho}\PR{\brho}\\
\label{phi.p}
    &\propto\sum_{\brho \in \mfP_{\mcI}} p^{T-S(\brho)}(1-p)^{S(\brho)}\\
    &=\phi(p),
\end{align}
where $\pdf(p)$ is the prior distribution of the parameter $p$ and we assume a uniform distribution, \ie  $\pdf(p)=2, p\in[0.5,1]$ which explains \eqref{likelihood.W.p}. 

The difference with the approaches described in \secref{Sec:property.opt.solution} appears in \eqref{Pr.W.rho.p.marginalization}: unlike \cite{Anderson19} or \cite{Cameron21} which assign a particular role to the optimal solutions $\hat\brho\in\hat\mcR$, we treat the order $\brho$ as a random variable and marginalize over it, assuming that it is uniformly distributed over the set of permutations $\mfP_{\mcI}$ (this is done in \eqref{phi.p}).

%%%%%%%%%%%%%%%%%%%%%%%%%%%%%%%%%%%%%%%%%%%%%%%%%%%%%%
\subsection{Slater spectrum}\label{Sec:Spectrum}

The explicit order enumeration in \eqref{phi.p} has a complexity of $M!$ that is prohibitive even for moderate $M$. Our goal is to reduce complexity by exploiting the structure of solutions.

First, we note that the number of comparisons that are inconsistent with the order $\brho$ is obviously a non-negative integer that cannot be greater than $T$, \ie $S(\brho)\in\set{0,\ld,T}$. It is possible to rewrite \eqref{phi.p} as follows:
\begin{align}
\label{phi.p.spectrum}
    \phi(p)
    &=
    \sum_{t=0}^T a_t p^{T-t}(1-p)^{t}
\end{align}
where
\begin{align}
\label{a.t}
    a_t &=  \sum_{\brho\in\mfP_{\mcI}} \IND{S(\brho)=t},\quad t=0,\ld, T
\end{align}
is the number of orders $\brho$ (in the set $\mfP_{\mcI}$) with the inconsistency index equal to $S(\brho)=t$. The sequence $\set{a_t}_{t=0}^T$ will be called \emph{the Slater spectrum} and, by construction, satisfies the conditions: $\sum_{t=0}^T a_t = M!$. Moreover, for any order $\brho$, it is enough to reverse it (\ie enumerate its elements backward) and, for the obtained order $\brho'=[\rho_M,\rho_{M-1}, \ld, \rho_1]$, all disagreements  with the order are transformed into agreements, and vice versa, \ie $S(\brho)=T-S(\brho')$, $a_{T-t}=a_t$, and, using \eqref{Consistency.brho} yields
\begin{align}
\label{a.t.consistency}
    a_t &=  \sum_{\brho\in\mfP_{\mcI}} \IND{C(\brho)=t},\quad t=0,\ld, T;
\end{align}
in other words, the Slater spectrum and consistency spectrum are the same, the choice of using \eqref{a.t} or \eqref{a.t.consistency} is a matter of convenience, and the name is given due to historical reasons.

Due to \eqref{ML.definition.final.C.hat} and the symmetry we also have
\begin{align}
\label{spectrum.has.limited.support}
    t<\hat{S}\quad  \tr{or} \quad t>T-\hat{S} \implies a_t\equiv 0,
\end{align}
and we can rewrite \eqref{phi.p.spectrum} as
\begin{align}
    \label{phi.p.spectrum.truncated}
    \phi(p) = \sum_{t=\hat{S}}^{T-\hat{S}}a_t p^{T-t} (1-p)^t
\end{align}
that is, the first non-zero element $a_{t}$ is indexed with $t=\hat{S}$ (and the last non-zero element is indexed with $t=T-\hat{S}$).

The problem is now reduced to finding $\set{a_t}$ and we show in \secref{Sec:Recursive.Calculation} the algorithm which avoids explicit enumeration over $\mfP_{\mcI}$ which appear in \eqref{a.t.consistency}.

With the Slater spectrum known, we can easily find the posterior \gls{pdf} by normalizing the function $\phi(p)$ as follows:
\begin{align}
\label{posterior.normalized}
    \pdf(p|\matW)&=\frac{1}{Z}\phi(p),
\end{align}
where the normalization factor is given by
\footnote{
From the property of the beta distribution
\begin{align}
    \int_{0.5}^1 p^{T-t}(1-p)^t\dd p = \frac{I_{0.5}(t+1,T-t+1)}{(T+1)\nchoosek{T}{t}},
\end{align}
where  $I_x(\alpha,\beta)=\nchoosek{\alpha+\beta-2}{\alpha-1}(\alpha+\beta-1)\int_{0}^x p^{\alpha-1}(1-p)^{\beta-1}\dd p $ is the regularized incomplete beta function, the normalization factor in \eqref{integral.Z} is calculated as
\begin{align}
    Z&=\int_{0.5}^1 \phi(p) \dd p
    =
    \sum_{t=\hat{S}}^{T-\hat{S}} a_t \int_{0.5}^1 p^{T-t}(1-p)^{t} \dd p\\
    &=
    \frac{1}{T+1}\sum_{t=\hat{S}}^{T-\hat{S}} \frac{a_tI_{0.5}(t+1,T-t+1)}{\nchoosek{T}{t}}\\
\label{Z.final.line}
    &=
    \frac{0.5}{T+1}\sum_{t=\hat{S}}^{T-\hat{S}} \frac{a_t I_{0.5}(t+1,T-t+1)}{\nchoosek{T}{t}}+\frac{a_{T-t}I_{0.5}(T-t+1,t+1)}{\nchoosek{T}{T-t}},
\end{align}
where, in \eqref{Z.final.line} we add the series enumerating them backwards (hence the multiplication by $0.5$). 

From the property $1-I_x(\alpha,\beta)=I_{1-x}(\beta,\alpha)$, we obtain $I_{0.5}(\alpha,\beta)+I_{0.5}(\beta,\alpha)=1$ which, plugged in \eqref{Z.final.line}, yields \eqref{integral.Z} if we exploit the symmetry $a_t=a_{T-t}$.
}
\begin{align}
\label{integral.Z}
    Z &=\int_{0.5}^1 \phi(p) \dd p
    %=\frac{1}{T+1}\left[\sum_{t=\hat{S}}^{\lfloor{(T-1)/2}\rfloor} \frac{a_t}{\nchoosek{T}{t}} +\frac{a_{T/2}}{2\nchoosek{T}{T/2}}\IND{T\tr{~even}}\right];
    =\frac{0.5}{T+1}\sum_{t=\hat{S}}^{T-\hat{S}} \frac{a_t}{\nchoosek{T}{t}};
\end{align}

We easily see that \eqref{integral.Z} the terms under summation are identical for two indices $t$ and $T-t$. However, we keep this redundancy for the sake of simplified notation (when $T$ is even,  the symmetry breaks and the term for $a_{T/2}$ would require a separate notation).

%%%%%%%%%%%%%%%%%%%%%%%%%%%%%%%%%%%%%%%%
\subsection{Slater index vs Slater spectrum}\label{Sec:Approximations.Slater.spectrum}

The main conclusions from the above development are summarized in the following, which is obvious from the construction of the Slater spectrum.
\begin{proposition}[Sufficient/insufficient statistics]\label{Prop:Sufficient.Statistics}
~\\
    1. The Slater spectrum is the sufficient statistic for $p$. \\
    2. The Slater index $\hat{S}$ is \emph{not} a sufficient statistic for $p$. % The exception is a ``degenerate" spectrum where $a_{t}\equiv 0, \forall t\neq \hat{S}$.  
\end{proposition}

Consequently, we may use the Slater spectrum in standard probabilistic and statistical tools to draw conclusions about $p$ without losing any information that can be provided by the matrix $\matW$.

Conversely, using only the Slater index $\hat{S}$, conclusions about $p$ (thus, also conclusions about rankability) will be approximate. The corollary conclusion is that the methods described in \secref{Sec:Slaters.index} and \secref{Sec:property.opt.solution} are approximate.

However, although relying only on the Slater index must lead to approximate conclusions, this is not to say that it is not useful. Indeed, it is better to be approximately correct than to not be able to say anything exactly, which may be the case if the calculation of the Slater spectrum is too complex to be practical.

Moreover, since the Slater index (or its transformation, \eg \eqref{degree.linearity}) has often been used in the literature, \eg\cite{Slater61}, \cite{Charon10}, \cite{Anderson19}, it is interesting to look at its role. 

From the perspective offered by \propref{Prop:Sufficient.Statistics}, using only the Slater index to estimate $p$, may be seen as using a ``degenerate" form of the Slater spectrum 
\begin{align}
    \label{degenerate.spectrum}
    \tilde{a}_t = a_t \IND{t=\hat{S} \vee t=T-\hat{S}},
\end{align}
and the corresponding ``degenerate" posterior \gls{pdf} is obtained by replacing $a_t$ with $\tilde{a}_t$ in \eqref{phi.p.spectrum.truncated}
\begin{align}
\label{tilde.pdf}
    \tilde\pdf(p|\matW)&=\frac{1}{\tilde{Z}}\tilde\phi(p)\\
\label{tilde.phi}
    \tilde\phi(p)
    &=
    a_{\hat{S}}
    \left[p^{T-\hat{S}}(1-p)^{\hat{S}} 
    +
    p^{\hat{S}}(1-p)^{T-\hat{S}}
    \right]\\
\label{tilde.Z}
    \tilde{Z}&=
    \frac{a_{\hat{S}}}{(T+1)\nchoosek{T}{\hat{S}}}
\end{align}
where, we immediately see that, due to normalization by $\tilde{Z}$, we might ignore the value of the coefficient $a_{\hat{S}}$. This would mean that knowing the number of optimal solutions $a_{\hat{S}}=|\hat{\mcR}|$ does not help to determine the rankability of $\matW$ under the assumed spectrum form \eqref{degenerate.spectrum}.

Thanks to \eqref{tilde.pdf}, we may work in the domain of the posterior distributions $\pdf(p|\matW)$ and this may provide us with an intuitive understanding of the conditions under which the approximations based on the Slater index, that is, using the degenerate spectrum $\set{\tilde{a}_t}$ instead of $\set{a_t}$, may be useful.

We do it considering two limiting cases. First, we assume that $\pdf(p|\matW)$ is ``concentrated" in the vicinity of $p=0.5$, \eg $\frac{1}{Z}\int_{0.5}^{0.5+\epsilon} \pdf(p|\matW) \dd p \approx 1$, where $\epsilon$ is a small number. Intuitively, this would mean that the data is not rankable. In the second case, we assume that$\pdf(p|\matW)$ is concentrated close to $p=1$, $\frac{1}{Z}\int_{1-\epsilon}^{1} \pdf(p|\matW) \dd p \approx 1$ which would be a good indicator of rankability of $\matW$.

To deal with the second case, changing variable $p=1-\epsilon$, with inconsequential abuse of notation, we write \eqref{phi.p.spectrum.truncated} as a function of $\epsilon$
\begin{align}
\label{approximate.p=1}
    \phi(\epsilon)
    &=
    \sum_{\hat{S}}^{T-\hat{S}}
    a_t(1-\epsilon)^{T-t}\epsilon^{t}
    \approx
    a_{\hat{S}}(1-\epsilon)^{T-\hat{S}}\epsilon^{\hat{S}}\\
\label{approximate.p=1.b}
    &
    \approx \tilde\phi(\epsilon)
\end{align}
where the approximation in \eqref{approximate.p=1} is valid if $\epsilon^{t+1}a_{t+1}\ll\epsilon^{t}a_t$ (which will hold for sufficiently small $\epsilon$). The same approximation applied to \eqref{tilde.phi} allows us to write \eqref{approximate.p=1.b} which essentially means that the Slater's index is sufficient to describe the posterior pdf $\pdf(p|\matW)$ for sufficiently small $\epsilon$. Actually, in this case, we say that the entire posterior \gls{pdf} may be approximated using only one term of $\phi(p)$, \ie
\begin{align}
\label{one.term.pdf}
    \pdf(p|\matW) \propto p^{T-\hat{S}}(1-p)^{\hat{S}}. 
\end{align}

Similarly, to deal with the case when the \gls{pdf} is concentrated in the vicinity of $p=0.5$, after the change of variable $p=0.5+\epsilon$, we can write \eqref{phi.p.spectrum.truncated} as
\begin{align}
\label{approximate.p=0.5}
    \phi(\epsilon)
    &=
    (0.5-\epsilon)^T\sum_{\hat{S}}^{T-\hat{S}}
    a_t\left(\frac{0.5+\epsilon}{0.5-\epsilon}\right)^t.
\end{align}
In this case, for small $\epsilon$, the entire spectrum $\set{a_t}$ is required to define $\phi(\epsilon)$, which, therefore, cannot be well approximated using only the Slater index.

Therefore, if $\matW$ is not rankable (the \gls{pdf} concentrated in the vicinity of $p=0.5$), the Slater index is not sufficient to characterize the posterior \gls{pdf}, while, if the data is rankable, the Slater index can, indeed, be used to describe well the most significant part of the posterior pdf $\pdf(p|\matW)$. %While it is still true that the form of the \gls{pdf} cannot be well characterized in the vicinity of $p=0.5$, by assuming that the \gls{pdf} is concentrated this part of the \gls{pdf} is negle.

However, leveraging this observation is not immediately obvious because, to know if we can rely on the Slater index alone, we need to know if the data is rankable beforehand which, of course, defeats the purpose of using the Slater index to analyze the rankability.

Visualizing the relationship between the degenerate and the true distributions may be difficult and, in practice, it is preferable to use parameters of the posterior \gls{pdf} $\pdf(p|\matW)$ such as the mode or the posterior mean of $p$.

\subsection{Mean}\label{Sec:Posterior.Mean}
A simple way to characterize the posterior distribution $\pdf(p|\matW)$ is through the posterior mean
\begin{align}
    \Ex[p] & = \int_{0.5}^1 p\pdf(p|\matW)\dd p\\
    &=
    \frac{1}{Z}\sum_{t=\hat{S}}^{T-\hat{S}} 
    a_t\int_{0.5}^1 p^{T-t+1}(1-p)^t\dd p\\
    &=\frac{1}{Z(T+2)}
    \sum_{t=\hat{S}}^{T-\hat{S}} 
    \frac{a_t I_{0.5}(t+1,T-t+2)}{\nchoosek{T+1}{t}},
\end{align}
where %$I_x(\alpha,\beta)=\nchoosek{\alpha+\beta-2}{\alpha-1}(\alpha+\beta-1)\int_{0}^x p^{\alpha-1}(1-p)^{\beta-1}\dd p $ is 
the regularized incomplete beta function $I_x(\alpha,\beta)$ is available in popular numerical packages.

The degenerate posterior mean is obtained using degenerate Slater spectrum \eqref{degenerate.spectrum}
\begin{align}
    \Ex[\tilde{p}]
    &=\frac{a_{\hat{S}}}{\tilde{Z}(T+2)}
    \sum_{t\in\set{\hat{S},T-\hat{S}}} 
    \frac{ I_{0.5}(t+1,T-t+2)}{\nchoosek{T+1}{t}}\\
    &=
    \frac{(T+1)\nchoosek{T}{\hat{S}}}{(T+2)}
    \sum_{t\in\set{\hat{S},T-\hat{S}}} 
    \frac{ I_{0.5}(t+1,T-t+2)}{\nchoosek{T+1}{t}}.
\end{align}

Again, we keep the formulas simple at the expense of some redundancy. 

%%%%%%%%%%%
\subsection{Mode}\label{Sec:Mode}
The posterior distribution \eqref{posterior.normalized} can be also characterized by its mode
\begin{align}
\label{p.MAP}
    \check{p} & = \argmax_{p\in[0.5,1]}\phi(p).
\end{align}
%which is the same as the \gls{ml} estimate of $p$.\footnote{Note that \eqref{p.MAP} is not the same as \eqref{joint.ML} because in $\phi(p)$ in \eqref{p.MAP} is obtained by marginalization over $\brho$, while \eqref{joint.ML} relies on the maximization over $\brho$.}

We may then use it in a way somewhat similar to what is done in Slater's methods: if the most likely explanation of the matrix $\matW$ is $\check{p}=0.5$, we may declare $\matW$ not rankable. However, unlike in Slater's method, we do not test the binary hypothesis, and obtaining $\check{p}\approx 0.5$ also may be treated as evidence for non-rankability.

To comment on the location of the mode, let us calculate the first and the second derivatives of $\phi(p)$:
\begin{align}
    \dot\phi(p)
    % &=
    % \sum_{t=0}^{T} a_t p^{t}(1-p)^{T-t} \left[\frac{t}{p}- \frac{T-t}{1-p}\right]\\
    &=
    \sum_{t=0}^{T} a_t p^{t}(1-p)^{T-t} \frac{t-pT}{p(1-p)}\\
\label{ddot.phi.p}
    \ddot\phi(p)
    % &=
    % \sum_{t=0}^{T}
    % \left( 
    % a_t p^{t}(1-p)^{T-t} \frac{(t-pT)^2 + [-Tp(1-p)-(t-pT)(1-2p)]}{p^2(1-p)^2}
    % \right)\\
    % &=
    % \sum_{t=0}^{T}
    % \left( 
    % a_t p^{t}(1-p)^{T-t} \frac{(t-pT)^2 + [-Tp+Tp^2-t(1-2p)+Tp(1-2p)]}{p^2(1-p)^2}
    % \right)\\
    &=
    \sum_{t=0}^{T}
    a_t p^{t}(1-p)^{T-t} \frac{(t-pT)^2 + [t(2p-1)-Tp^2]}{p^2(1-p)^2}.
\end{align}

From the symmetry of the spectrum $a_{T-t}=a_t$, we easily obtain the following:
\begin{align}
    % \sum_{t=0}^T a_t t =   \sum_{t=0}^T a_{T-t} t\\
    \sum_{t=0}^T a_t t =   \frac{T}{2}\sum_{t=0}^T a_t
\end{align}
which allows us to calculate the first derivative at $p=0.5$ as
\begin{align}\label{dot.phi=0}
    \dot\phi(0.5)
    &=
    0.5^{T-2}\sum_{t=0}^T a_t (t-0.5T)=0;
\end{align}
this means that the function $\phi(p)$ has an extremum at $p=0.5$ and we would like to verify if it is a maximum or the minimum of $\phi(p)$. 
 
The results we observed are similar in simulations and in empirical data, \cf \figref{Fig:pdf_0.6}, \figref{Fig:phi_func_real}, lead to the following:
\begin{conjecture}[Maximum of $\phi(p)$]\label{Conjecture:phi}
The function $\phi(p)$ has only one maximum on the interval $[0.5, 1)$.
\end{conjecture}

We can express now the following sufficient conditions for determining the location of the mode:
\begin{align}
\sigma &= \ddot\phi(0.5)\\
\label{sigma.positive}
    \sigma > 0 &\implies \check{p}>0.5\\
\label{sigma.negative}
    \sigma < 0 &\implies \check{p}=0.5,
\end{align}
where Conjecture~\ref{Conjecture:phi} is necessary to guarantee that, if $p=0.5$ is a local maximum, it is also the global one; this is expressed in \eqref{sigma.negative}. Then, the most likely explanation for the matrix $\matW$ is the model with $\check{p}=0.5$, and we may declare $\matW$ not rankable. On the other hand, when $\sigma>0$ we know that $p=0.5$ is not a maximum and we may declare $\matW$ rankable even without Conjecture~\ref{Conjecture:phi}.

Note that the conditions in \eqref{sigma.positive}-\eqref{sigma.negative} are only sufficient because, if $\sigma=0$, the ambiguity about $p=0.5$ being minimum or maximum cannot be resolved and we need higher-order derivatives of $\phi(p)$, which makes the derivations slightly more involved.

To find the mode $\check{p}$, we may proceed in two steps. First, we verify if the conditions \eqref{sigma.negative} are satisfied, as then we immediately obtain $\check{p}=0.5$. If, on the other hand, \eqref{sigma.positive} is satisfied, we need to numerically solve equation $\dot\phi(p)=0, p\in(0.5,1)$.

Although finding the mode $\check{p}$ can be done numerically, the main point of the above analysis is to find conditions for the rankability \emph{without} solving numerical equations.

This is possible if we use the degenerate spectrum $\tilde{a}_t$ from \eqref{degenerate.spectrum}: we find $\sigma$ to be a quadratic function of the Slater index $\hat{S}$
\begin{align}
    \tilde{\sigma} = \ddot{\tilde{\phi}}(0.5) = 
    \frac{a_{\hat{S}}}{2^{T-4}}\Big[ (\hat{S}-0.5T)^2 -T0.25\Big]
\end{align}
so, \eqref{sigma.positive}-\eqref{sigma.negative} become
\begin{align}
\label{convexity.simple.hat.S}
    \tilde{\sigma}>0 &\iff \hat{S}<\hat{S}_{\tr{th}},\\
\label{concavity.simple.hat.S}
    \tilde{\sigma}<0 &\iff \hat{S}>\hat{S}_{\tr{th}},\\
    \hat{S}_{\tr{th}}&=0.5(T-\sqrt{T})
\end{align}

\textbf{Relationship to the degree of linearity} %$\lambda$, \eqref{degree.linearity}}

Using \eqref{degree.linearity}, the relationships \eqref{convexity.simple.hat.S}-\eqref{concavity.simple.hat.S} may be expressed as
\begin{align}
    \tilde{\sigma}>0 &\iff \lambda>\lambda_{\tr{th}},\\
    \tilde{\sigma}<0 &\iff \lambda<\lambda_{\tr{th}},\\
\label{lambda.threshold}
    \lambda_{\tr{th}}&=\frac{1}{2}(1+\frac{1}{\sqrt{T}}).
\end{align}
 
However, it seems that in most practical cases, condition $\lambda>\lambda_{\tr{th}}$ will be easily satisfied, especially for large $T$ (\eg for $T=100$, we have $\lambda_{\tr{th}}=0.55$). Therefore, although $\lambda_{\tr{th}}$ provides a quantitative condition for estimating rankability using only the degree of linearity $\lambda$, this condition is very weak, which is the consequence of using the degenerate Slater spectrum $\set{\tilde{a}_t}$. 

Assuming that the posterior \gls{pdf} is concentrated around $p=1$, \eqref{one.term.pdf} holds and is identical to the term under optimization in \eqref{joint.ML.definition}. Therefore, 
\begin{align}
    \check{p}\approx \lambda.
\end{align}
In other words, if \eqref{one.term.pdf} holds, the degree of linearity, $\lambda$, can be treated as a mode of posterior distribution.

\subsection{Inference from multiple observations}\label{Sec:Multipl.Realizations}

Let us assume that we deal with $L$ comparisons of different objects; \ie we observe the matrices $\set{\matW_l}_{l=1}^L$, which depend on orders $\brho_l, l=1,\ld,L$, where $\brho_k\neq\brho_l, l\neq k$. In this case, if we want to find the rankings $\hat\brho_l$, we deal simply with multiple instances of the \gls{lop}.

However, if we assume that the observation matrices $\matW_l$, conditioned on the orders $\brho_l$, are generated according to the model \eqref{Pr=p}-\eqref{Pr=q}, and that the orders $\brho_l$ are independent, then the probabilistic approach we propose confers us another advantage because we can use all the matrices to obtain the posterior \gls{pdf} of $p$
\begin{align}
\label{product.phi}
    \pdf(p|\set{\matW_l}_{l=1}^L) 
    & \propto 
    \PR{\matW_1,\ld,\matW_L|p}\propto\prod_{l=1}^L\phi(p|\matW_l).
\end{align}

Of course, multiple comparisons of \emph{different} objects may share the same model \eqref{Pr=p}-\eqref{Pr=q}. Such an assumption may be justified if the comparisons are done on the same ``type" of objects and the results are affected by the same uncertainty. For example, in sport leagues, the comparisons between the teams (made through games) are made according to the same rules. Also, in the sport context, the assumption of independence of orders $\brho_l$ (between seasons) can be justified if the levels of the teams are equalized through the annual exchange of players.

Since $\phi(p|\set{\matW_l}_{l=1}^L)$ is a product of functions, finding its mean and/or mode analytically seems difficult, but is done straightforwardly numerically.

If, on the other hand, we use the approximation \eqref{one.term.pdf}, it is easy to see that \eqref{product.phi} may be expressed as
\begin{align}
    \pdf(p|\set{\matW_l}_{l=1}^L)
    &\propto \prod_{l=1}^L p^{T_l-\hat{S}_l}(1-p)^{\hat{S}_l}\\
    &= p^{L\ov{T}-L\ov{\hat{S}})}(1-p)^{L\ov{\hat{S}}},
\end{align}
where $\ov{\hat{S}}=\frac{1}{L}\sum_{l=1}^L\hat{S}_l$ is the average Slater index, and $\ov{T}=\frac{1}{L}T_l$ is average number of observations. Here, $\hat{S}_l$ and $T_l$ are, respectively, the Slater index and the number of observations in $\matW_l$ (that is, we recognize that problems may have different structure); for example, in sports, it may happen that some seasons have games canceled or number of teams increased.

By analogy to \eqref{joint.ML.definition}, we conclude that the mode of the joint \gls{pdf} \eqref{product.phi} is then given by
\begin{align}
\label{degree.linearity.joint}
    \check{p} &\approx \lambda_{\tr{joint}} = 1- \frac{\ov{\hat{S}}}{\ov{T}};
\end{align}
in other words, the ratio of the average Slater index and the average number of observations should be treated as the ``degree of linearity" index for the joint estimation. Note that, $\lambda_{\tr{joint}}$ is \emph{not} the same as the average of individual linearity indices $\lambda_l=1-\frac{\hat{S}_l}{T_l}$. The equivalence appears only if all $T_l$ are equal. 

%%%%%%%%%%%%%%%%%%%%%%%%%%%%%%%%%%%%%%%%%%%%%%%%%%%%%%%%%%%%
\section{Algorithms}\label{Sec:Recursive.Calculation}

In order to evaluate the rankability criteria shown in \secref{Sec:Posterior.Mean} or \secref{Sec:Mode}, we may need to obtain the spectrum $\set{a_t}$. We show how to do it efficiently, by exploiting the structure of the problem.

\subsection{Recursive calculation of Slater spectrum}\label{Sec:Recursive.Slater.spectrum}
For any subset $\mcI'\subset \mcI$, let $\mfP_{\mcI'}$ be the space of permutation of elements in $\mcI'$, and $C(\brho')$ -- the  consistency index of the order $\brho'\in\mfP_{\mcI'}$, where, taking into account the cardinality of $\mcI'$,  we redefine \eqref{Consistency.brho} as follows:
\begin{align}
    \label{C.brho.redefined}
    C(\brho') &= \sum_{m=1}^{|\mcI'|}\sum_{n=m+1}^{|\mcI'|} w_{\rho'_m,\rho'_n}.
\end{align}

Let us define a mapping from the order $\brho'$ onto a polynomial in an auxiliary variable $u$
\begin{align}
\label{polynomial.definition}
    \brho' \mapsto u^{C(\brho')},
\end{align}
and, since the degree of the polynomial in \eqref{polynomial.definition} gives us the consistency index of the order $\brho'$, then 
\begin{align}
\label{spectrum.I}
    G(u; \mcI') 
    &= \sum_{\brho'\in \mfP_{\mcI'}}u^{C(\brho')}=\sum_{t=0}^Ta_{\mcI',t}u^t
\end{align}
is a polynomial whose coefficients $a_{\mcI',t}$ have meaning of the Slater spectrum and tell us how many orders $\brho'$ with consistency index $C(\brho')=t$ there are in the set of permutations $\mfP_{\mcI'}$. We call $G(u;\mcI')$ the Slater spectrum (in the polynomial domain). In particular, for $\mcI=\mcI'$, the coefficients $a_{\mcI',t}$ define the Slater spectrum for the entire set $\mcI$. Therefore, the idea is to use the relationship between the spectra of the subsets and, recursively, arrive at the spectrum $G(u;\mcI)$. 

Let $\mcI''=\mcI'\backslash e$ be a subset of $\mcI'$, obtained by removing the element $e$ from $\mcI'$ (thus, $\mcI''$ has $|\mcI'|-1$ elements). Then, any order $\brho'\in \mfP_{\mcI'}$ can be represented as a concatenation of an element $\rho'_1=e$, with an order $\brho''\in\mfP_{\mcI'\backslash e}$, \ie  $\brho'=[e,\brho'']$. Thus, the summation over all the orders $\brho'\in\mfP_\mcI$, required to calculate \eqref{spectrum.I}, can be written as
\begin{align}
\label{smart.enumeration}
    \sum_{\brho'\in\mfP_{\mcI'}} u^{C(\brho')}
    &= 
    \sum_{e\in \mcI'}
    \sum_{\brho''\in\mfP_{\mcI'\backslash e}}
    u^{C([e,\brho''])}.
\end{align}

The consistency index of the order $\brho'=[e,\brho'']$ in \eqref{smart.enumeration} is calculated as follows:
\begin{align}
    C([e,\brho''])
    &=
    \sum_{m=1}^{|\mcI'|}\sum_{n=m+1}^{|\mcI'|}w_{\rho'_m,\rho'_n}\\
    &=
    \sum_{n=1}^{|\mcI'|-1}w_{e,\rho'_n}
    + \sum_{m=1}^{|\mcI'|-1}\sum_{n=m+1}^{|\mcI'|-1}w_{\rho''_m,\rho''_n}\\
\label{C.rho.prim.bis}
    &=
    d_e(\mcI') + C(\brho''),
\end{align}
where
\begin{align}
\label{C.cross.spectrum}
    d_e(\mcI')
    &=
    \sum_{n\in\mcI'} w_{e,n}.
\end{align}

We can now rewrite \eqref{spectrum.I}  using \eqref{smart.enumeration} and \eqref{C.rho.prim.bis}
\begin{align}
    G(u; \mcI') 
\label{spectrum.1}
    &=
    \sum_{e\in\mcI'}
    u^{d_e(\mcI')}
    \sum_{\brho''\in\mfP_{\mcI'\backslash e}}
    u^{C(\brho'')}\\
\label{spectrum.2}
    &=
    \sum_{e\in\mcI'}
    u^{d_e(\mcI')}G(u;\mcI'\backslash e),
\end{align}
where \eqref{spectrum.2} uses the definition of the spectrum from \eqref{spectrum.I} that is valid for any subset of $\mcI$. %Clearly, to calculate the spectrum $G(u;\mcI')$ we need to know the spectra $G(u;\mcI\backslash{e})$ of all subsets $\mcI\backslash{e}, e\in\mcI$. 

To generalize the above approach, let $\mcF_k=\set{\mcI': \mcI'\subset\mcI, |\mcI'|=k}$ denote the family of all distinct subsets of $\mcI$ that have $k$ elements.\footnote{A family defined in this way is a set, but we use a different name to make a distinction between its elements, which are themselves the sets.} Obviously, there are $|\mcF_k|=\nchoosek{M}{k}$ subsets in  a family $\mcF_k$. In particular, there is only one subset with $M$ elements -- the set $\mcI$ itself, thus $\mcF_M = \set{\mcI}$. Similarly, there are $M$ one-element subsets in $\mcF_1 = \set{\{1\},\{2\},\ld,\{M\}}$.

Applying \eqref{spectrum.1}, to calculate the spectrum $G(u;\mcI)$  we need to know the spectra $G(u;\mcI\backslash{e})$ for all subsets $\mcI'=\mcI \backslash{e} \in\mcF_{M-1}$. Next, to calculate the spectrum $G(u;\mcI')$, we need the spectra $G(u;\mcI'')$ for all subsets $\mcI''=\mcI'\backslash{e}\in\mcF_{M-2}$, etc. This leads to the recursion described with pseudo-code shown in Algorithm~\ref{Algo:Slater.Spectrum}.

We initialize the algorithm by calculating the spectra for all subsets composed of one element $\mcI'\in\mcF_1$; this is trivial: $G(u;\mcI')\equiv u^0 = 1$ and means that for one element, no comparison is possible, so the consistency index is equal to zero.

Next, we apply \eqref{spectrum.2} for each $\mcI'\in\mcF_{k}$ for $k=2,3,\ld, M$, where the final step $k=M$ yields the spectrum $G(u;\mcI)$.

% ------------------------------- algorithm -----------------------
\begin{algorithm}
    \caption{Recursive calculation of the Slater spectrum} \label{Algo:Slater.Spectrum}
\begin{algorithmic}[1]
\State Input: $\matW\in\Natural^{M\times M}$ \Comment Matrix of the results
\State Initialization:
\State $\mcF_1=\set{\{1\},\{2\},\ld, \{M\}}$\label{DEF.F.1} \Comment all subsets of $\mcI$ with one element
\For{$\mcI'\in \mcF_1$}\label{FOR.I.prim.1}
    \State $G(u;\mcI')\leftarrow 1$\Comment $G(u;\mcI')=u^0$
\EndFor
\State Recursion:
\For{ $k = 2, \ld, M$ }\label{FOR.k}
    \State Create $\mcF_k$ \Comment all subsets of $\mcI$ with $k$ elements, $|\mcF_k|=\nchoosek{M}{k}$
    \For{$\mcI' \in\mcF_k$}\label{FOR.mcI.prim}
        \State $G(u;\mcI')\leftarrow 0$
        \For{$e\in\mcI'$}    \label{FOR.i}
        % \State $c \leftarrow $\Comment use \eqref{C.cross.spectrum}
	\State $G(u; \mcI') \leftarrow  G(u; \mcI') + u^{d_e(\mcI')} G\big(u;\mcI'\backslash e\big)$ \label{LINE.polynomial}
        \EndFor
    \EndFor
\EndFor % i

\State Output: $G(u; \mcI)=G(u;\mcI'), \mcI'\in\mcF_M$
\end{algorithmic}
\end{algorithm}
% ------------------------------- end algorithm -------------------

Some implementation comments are in order.
\begin{enumerate}
    \item The spectra are indexed with sets $\mcI'$, that is, the ordering of elements in $\mcI'$ does not matter. For example, the spectrum $G(u;\set{1,2})$ is exactly the same as $G(u;\set{2,1})$. Since we do not want to store the same data with different indices, in practice, it is preferable to order the elements in $\mcI'$, \eg lexicographically, so that a subset obtained by removing one element $e$, \ie $\mcI''=\mcI'\backslash e$, becomes a valid index to the spectrum $G(u;\mcI'')$. 
    \item The spectrum $G(u;\mcI')$ can be represented as vectors $\ba_{\mcI'}=[a_{\mcI',0},\ld, a_{\mcI',T}]$ so the multiplication by $u^{d_e(\mcI)}$ in \eqref{spectrum.2} (see line~\ref{LINE.polynomial} of Algorithm~\ref{Algo:Slater.Spectrum}) amounts to increasing the degree of all terms in $G(u;\mcI'\backslash e)$ by $d_e(\mcI')$. That is, before being added, all coefficients must be right-shifted as follows:
    \begin{align}
        a_{\mcI',k} = \sum_{e\in\mcI'} a_{\mcI'\backslash e, k-d_e(\mcI')},\quad k=0,\ld,T,
    \end{align}
    where, for convenience, we set $a_{\mcI',k}\equiv 0$ for $k<0$.
    \item It is possible to obtain \eqref{C.cross.spectrum} recursively, saving calculation. This is done as follows for any $\mcI'=\set{e'_1,\ld,e'_k}\in\mcF_k$ 
    \begin{align}
    \label{de1}
        d_{e'_1}(\mcI') & = d_{e'_1}(\mcI'\backslash e'_2) + w_{e'_1,e'_2}\\
    \label{dej}
        d_{e'_j}(\mcI') & = d_{e'_j}(\mcI'\backslash e'_1) + w_{e'_j,e'_1}, \quad j=2,\ld,k.
    \end{align}

    This simplification is interesting for comparison purposes; however, it is not certain that significant gains materialize in practice. For example, in our implementation (in Python), significant complexity is associated with the indexing strategy. Thus, implementing \eqref{de1}-\eqref{dej} was immaterial for the execution time and the codes we provide in \github calculate $d_e(\mcI')$ directly using \eqref{C.cross.spectrum} as it simplifies the implementation and eases understanding of the algorithms.
    
\end{enumerate}

\begin{example}[Finding the Slater spectrum]\label{Example:Slater.spectrum}
Let us consider the case of $M=4$, where each pair of objects is compared once, \ie $T_{m,n}=1, \forall m,n$, and consider the following observation matrix:
\begin{align}
    \matW
    &=
    \left[
    \begin{array}{cccc}
    0 & 1 & 1 & 0 \\
    0 & 0 & 1 & 1 \\
    0 & 0 & 0 & 0 \\
    1 & 0 & 1 & 0
    \end{array}
    \right].
\end{align}

The steps of Algorithm~\ref{Algo:Slater.Spectrum} may be traced on a graph shown in \figref{fig:Algorithm}, where the nodes, indexed with the subsets $\mcI'$ (or, simply, ``nodes $\mcI'$'') shown as rectangles, contain the vectors of coefficients $a_{\mcI',k}, k=0,\ld, T$, where $T=6$ because we have six comparison results defined by $\matW$.

The nodes $\mcI'$ from the same family $\mcF_k$ are stacked in the same column. In particular, the initialization (loop in line \ref{FOR.I.prim.1} of Algorithm~\ref{Algo:Slater.Spectrum}) produces the spectra $G(u;\mcI')=1$ for all $\mcI'= \set{1}, \set{2}, \set{3}, \set{4}$ shown in the right-most column in \figref{fig:Algorithm}.

The edges connecting the nodes indicate the relationship between a node $\mcI'\in\mcF_{k}, k=2,3,4$ (on the left-end of the edge) and its ``parent'' node $\mcI'\backslash e\in\mcF_{k-1}, e\in\mcI'$ (on the righ-end of the edge). The edge is labeled with ``$e; d_e(\mcI')$''; see \eqref{C.cross.spectrum}.

The loop in line \ref{FOR.i} of Algorithm~\ref{Algo:Slater.Spectrum} starts by right-shifting each parent spectra vector $[a_{\mcI'\backslash e, 1},\ld, a_{\mcI'\backslash e, T}]$ by $d_e(\mcI')$, and then adds them element-wise. For example, it is easy to see that two edges entering any subset $\mcI'=\set{i,j}\in\mcF_2$ must be labeled with $d_{i}(\mcI')=0$ and $d_{j}(\mcI')=1$ or with $d_{i}(\mcI')=1$ and $d_{j}(\mcI')=0$; this is because, in our example, there is only one comparison made between any pair of objects. Since all the parent spectra are, by definition, equal to $G(u; \mcI')=1$ \ie characterized by a vector $[1,0,0,0,0,0]$, the spectra of all sets $\mcI'\in\mcF_2$ must be defined by a vector $[1,1,0,0,0,0]$, as we can see in the second right-most column in \figref{fig:Algorithm}.

The steps shown in \figref{fig:Algorithm} may be reproduced using the program shown in \github.

\end{example}

%%%%%%%%%%%%%%
\begin{figure}[bt]
\begin{center}
\scalebox{\sizf}{%!TEX root =  ../Phase.Track.HARQ.tex
\pgfdeclarelayer{background}
\pgfdeclarelayer{foreground}
\pgfsetlayers{background,main,foreground}

\tikzstyle{rect_my2} = [draw, rectangle, minimum width=2cm, text width=2.3cm, fill=gray!15, 
  text centered,  minimum height=.9cm]
%=================
% \begin{center}
\begin{tikzpicture}[trim left=0cm,node distance=3cm]

  %%%%%--------  Set nodes -------- %%%%%%%%%%%%%%%%  
  % F_4
\node[rect_my2]	(I_4) 		[label=north:{$\set{1,2,3,4}$}]	{$[0,3,6,6,6,3,0]$};
% F_3
\node[rect_my2] 	(I_3_1) 	[right of=I_4,node distance=4cm,yshift=3cm,label=north:{$\set{1,2,3}$}]	{$[1,2,2,1,0,0,0]$};
\node[rect_my2] 	(I_3_2) 	[below of=I_3_1,yshift=1cm,label=north:{$\set{1,2,4}$}]	{$[0,3,3,0,0,0,0]$};
\node[rect_my2] 	(I_3_3) 	[below of=I_3_2,yshift=1cm,label=north:{$\set{1,3,4}$}]	{$[1,2,2,1,0,0,0]$};
\node[rect_my2] 	(I_3_4) 	[below of=I_3_3,yshift=1cm,label=north:{$\set{2,3,4}$}]	{$[1,2,2,1,0,0,0]$};
% F_2
\node[rect_my2] 	(I_2_1) 	[right of=I_3_1,node distance=4cm, yshift=2cm,label=north:{$\set{1,2}$}]	{$[1,1,0,0,0,0,0]$};
\node[rect_my2] 	(I_2_2) 	[below of=I_2_1,yshift=1cm,label=north:{$\set{1,3}$}]	{$[1,1,0,0,0,0,0]$};
\node[rect_my2] 	(I_2_3) 	[below of=I_2_2,yshift=1cm,label=north:{$\set{1,4}$}]	{$[1,1,0,0,0,0,0]$};
\node[rect_my2] 	(I_2_4) 	[below of=I_2_3,yshift=1cm,label=north:{$\set{2,3}$}]	{$[1,1,0,0,0,0,0]$};
\node[rect_my2] 	(I_2_5) 	[below of=I_2_4,yshift=1cm,label=north:{$\set{2,4}$}]	{$[1,1,0,0,0,0,0]$};
\node[rect_my2] 	(I_2_6) 	[below of=I_2_5,yshift=1cm,label=north:{$\set{3,4}$}]	{$[1,1,0,0,0,0,0]$};
% F_1
\node[rect_my2] 	(I_1_1) 	[right of=I_3_1,node distance=8cm, label=north:{$\set{1}$}]	{$[1,0,0,0,0,0,0]$};
\node[rect_my2] 	(I_1_2) 	[below of=I_1_1,yshift=1cm,label=north:{$\set{2}$}]	{$[1,0,0,0,0,0,0]$};
\node[rect_my2] 	(I_1_3) 	[below of=I_1_2,yshift=1cm,label=north:{$\set{3}$}]	{$[1,0,0,0,0,0,0]$};
\node[rect_my2] 	(I_1_4) 	[below of=I_1_3,yshift=1cm,label=north:{$\set{4}$}]	{$[1,0,0,0,0,0,0]$};

 %%%%%--------    Draw Paths  -------- %%%%%%%%%%%%%%%%
 % F_1 to F_2
\path[line_my]   (I_1_1.west)	-- node[above, near end, sloped]{$2;0$}	(I_2_1.east);
\path[line_my]   (I_1_1.west)	-- node[above, near end, sloped]{$3;0$}	(I_2_2.east);
\path[line_my]   (I_1_1.west)	-- node[above, near end, sloped]{$4;1$}	(I_2_3.east);
\path[line_my]   (I_1_2.west)	-- node[below, near end, sloped]{$1;1$}	(I_2_1.east);
\path[line_my]   (I_1_2.west)	-- node[above, near end, sloped]{$3;0$}	(I_2_4.east);
\path[line_my]   (I_1_2.west)	-- node[above, near end, sloped]{$4;0$}	(I_2_5.east);
\path[line_my]   (I_1_3.west)	-- node[above, very near end, sloped]{$1;1$}	(I_2_2.east);
\path[line_my]   (I_1_3.west)	-- node[below, near end, sloped]{$2;1$}	(I_2_4.east);
\path[line_my]   (I_1_3.west)	-- node[above, near end, sloped]{$4;1$}	(I_2_6.east);
\path[line_my]   (I_1_4.west)	-- node[above, very near end, sloped]{$1;0$}	(I_2_3.east);
\path[line_my]   (I_1_4.west)	-- node[below, near end, sloped]{$2;1$}	(I_2_5.east);
\path[line_my]   (I_1_4.west)	-- node[below, near end, sloped]{$3;0$}	(I_2_6.east);
% F_2 to F_3
\path[line_my]   (I_2_1.west)	-- node[above, near start, sloped]{$3;0$}	(I_3_1.east);
\path[line_my]   (I_2_2.west)	-- node[above, near start, sloped]{$2;1$}	(I_3_1.east);
\path[line_my]   (I_2_4.west)	-- node[below, very near start, sloped]{$1;2$} (I_3_1.east);
\path[line_my]   (I_2_1.west)	-- node[below, near start, sloped]{$4;1$}	(I_3_2.east);
\path[line_my]   (I_2_3.west)	-- node[above, near start, sloped]{$2;1$}	(I_3_2.east);
\path[line_my]   (I_2_5.west)	-- node[below, very near start, sloped]{$1;1$} (I_3_2.east);
\path[line_my]   (I_2_2.west)	-- node[below, near start, sloped]{$4;2$}	(I_3_3.east);
\path[line_my]   (I_2_3.west)	-- node[below, near start, sloped]{$3;0$}	(I_3_3.east);
\path[line_my]   (I_2_6.west)	-- node[above, near start, sloped]{$1;1$} (I_3_3.east);
\path[line_my]   (I_2_4.west)	-- node[below, near start, sloped]{$4;1$}	(I_3_4.east);
\path[line_my]   (I_2_5.west)	-- node[below, near start, sloped]{$3;0$}	(I_3_4.east);
\path[line_my]   (I_2_6.west)	-- node[below, near start, sloped]{$2;2$} (I_3_4.east);
% F_3 to F_4
\path[line_my]   (I_3_1.west)	-- node[above, near start, sloped]{$4;2$}	(I_4.east);
\path[line_my]   (I_3_2.west)	-- node[above, near start, sloped]{$3;0$}	(I_4.east);
\path[line_my]   (I_3_3.west)	-- node[below, near start, sloped]{$2;2$}	(I_4.east);
\path[line_my]   (I_3_4.west)	-- node[below, near start, sloped]{$1;2$}	(I_4.east);

%%%% %% --- Draw Big Box
%\begin{pgfonlayer}{background}
%\node[bigbox_my,fill=blue!20, fit=(Dem1)(Demk)(Agg),label=north:$\DEMwo$]   (Dem_tot) {};
%\node[bigbox_my,fill=blue!20, fit=(Chan1)(Chank),label=north:Channel]   (Chan_tot) {};
%\node[below right] at (RX.north)  {Receiver};
%\end{pgfonlayer}

%%%%
\end{tikzpicture}
% \end{center}
}
\end{center}
\caption{Steps of Algorithm~\ref{Algo:Slater.Spectrum} applied to the matrix $\matW$ shown in \exref{Example:Slater.spectrum}. Nodes in the same column are indexed with the subsets $\mcI'\in\mcF_{k}$ (shown above the box); the right-most column shows the subsets from the family $\mcF_1$, and the left-most one shows $\mcI'=\mcI\in \mcF_4$. The vectors shown inside the nodes $\mcI'$ represent spectra $G(u;\mcI')$, \ie $[a_{\mcI',0},\ld,a_{\mcI',T}]$. The path connecting the subset $\mcI'\backslash e$ (on the right) to the subset $\mcI'$ (on the left) is labeled with ``$e;d_e(\mcI')$''.}
\label{fig:Algorithm}
\end{figure}
%%%%%%%%%%%%%%%

\textbf{Complexity evaluation}

The complexity of the Algorithm~\ref{Algo:Slater.Spectrum} is due to three nested loops (a) over $k=2,\ld, M$ (line \ref{FOR.k}), (b) over $\nchoosek{M}{k}$ elements of $\mcF_k$ (line \ref{FOR.mcI.prim}), and (c) over $k$ elements $e$ of $\mcI'$ (line \ref{FOR.i}). In the most inner loop, over $e$, we (i) calculate \eqref{C.cross.spectrum}, which requires a summation of $1$ elements of the matrix $\matW$ (using \eqref{de1}-\eqref{dej}), and (ii) add $k-1$ vectors (representing polynomials), each with $T$ elements. 

The total number of operations may be calculated as follows:
\begin{align}
\label{CMPLX.spectrum}
    \tr{CMPLX} 
    &= \sum_{k=2}^M \nchoosek{M}{k} \big[k+ T(k-1)\big]\\%= M(M-1)2^{M-2}\\
    &=O\Big(M^3 2^M \Big),%O\Big( (M^2 + T)2^M \Big),
\end{align}
where we use the relationship $\sum_{k=0}^M \nchoosek{M}{k}k=2^M M/2$ %$\sum_{k=2}^M \nchoosek{M}{k}k^2\sim 2^M M^2/4$, 
and assume that $T\sim M^2$, \eg if we make $K_{m,n}=K$ comparisons for each pair, then $T = KM(M-1)/2$.

For example, with $M=20$, we have the complexity $\tr{CMPLX}=O(10^{10})$ which is within reach of any modern computer, while the explicit enumeration in \eqref{phi.p} with the complexity of $O(M!) = 10^{18}$ -- is not.

The algorithm implemented in Python is available at \github.

An interesting feature of the algorithm is that even though we do not explicitly find the solution $\hat\brho$ to the ordering problem, we immediately know how many such solutions there are: this is the coefficient $a_{\mcI,\hat{S}}=a_{\mcI,T-\hat{S}}$, where $\hat{S}$ is the optimal Slater index in \eqref{ML.definition.final.C.hat}. Furthermore, with a minor modification of the above algorithm, and with much lower complexity, we can recover \emph{all}, $a_{\mcI,\hat{S}}$ solutions $\hat\brho$; this is shown below.

%%%%%%%%%%%%%%%%%%%%%%%%%%%%%%%%%%%%%%%%%%%%%%%%%%%%%%%
\subsection{Recovering all rankings $\hat\brho$}\label{Sec:All.Solutions}

Let $q(\mcI')=\max_{\brho\in\mcI'}\set{C(\brho)}$ be the maximal consistency index, \ie the degree of the spectrum polynomial $G(u;\mcI')$
\begin{align}
    q(\mcI') 
    &= \argmax_{t} \set{a_{\mcI',t} > 0}
\end{align}
which, from \eqref{spectrum.2} and \eqref{C.cross.spectrum}, is calculated as
\begin{align}
\label{recursive.order}
    q(\mcI') 
    &= \max_{e\in\mcI'} 
    \set{d_e(\mcI') + q(\mcI'\backslash e)},
\end{align}
and we are able to keep track of $q(\mcI')$ when executing the steps of the recursion \eqref{spectrum.2}, which, at the final step yields $q(\mcI) = T-\hat{S}$.

All optimal rankings may be written as $\hat\brho=[\hat{e}_1,\hat{\brho'}]$ for some $\hat{\brho'}\in\mcI\backslash{\hat{e}}_1$, and, from \eqref{C.rho.prim.bis}, $C(\hat\brho)=d_{\hat{e}_1}(\mcI) + C(\hat\brho')$. Furthermore, by definition, the consistency index of an optimal ranking is maximal, \ie $C(\hat\brho)=q(\mcI)$ which implies that the consistency index of $\hat\brho'\in\mcI\backslash\hat{e}_1$ is also maximal in $\mcI\backslash\hat{e}_1$, \ie $C(\hat\brho')=q(\mcI\backslash\hat{e}_1)$. 

Therefore, the element $e=\hat{e}_1$ which maximizes the expression in \eqref{recursive.order} and thus satisfies the following: 
\begin{align}
\label{order.index}
    q(\mcI\backslash \hat{e}_1)  = q(\mcI) - d_{\hat{e}_1}(\mcI),\quad \hat{e}_1\in\mcI,
\end{align}
is the first element of the optimal ranking $\hat\brho$. We say that node $\mcI\backslash{\hat{e}_1}$ is \emph{degree-compatible} with $\mcI$. Note that \eqref{order.index} may have many solutions $\hat{e}_1$, and therefore there may be more than one degree-compatible node $\mcI\backslash\hat{e}_1$.

Conversely, if $e\in\mcI$ does not maximize expression in \eqref{recursive.order} and therefore does not satisfy \eqref{order.index}, it means that $e$ is \emph{not} the first element of the ranking $\hat\brho$.

Therefore, although the spectrum $G(u;\mcI)$ does not give us direct information about the elements of the optimal rankings $\hat\brho$, this information may be retrieved from the edges connecting the node $\mcI$ to its parents $\mcI\backslash{e}$ and allows us to unveil a ranking associated with the node $\mcI\backslash\hat{e}_1$ 
\begin{align}
\label{tilde.brho.M-1}
 \hat\mcR_{\mcI\backslash\hat{e}_1}=\set{[\hat{e}_1,\cd,\ld,\cd]},
\end{align}
which is ``preliminary" in the sense that the elements $\rho_2,\ld,\rho_M$ are yet unknown as indicated by ``$\cd$'' placed at $M-1$ corresponding positions. %On the other hand, we know that at least one optimal ranking $\hat\brho$ will have its first element given by $\hat{\rho}_1=\hat{e}_1$.

In fact, we may say that the preliminary ranking associated with $\mcI$ is $\hat{\mcR}_{\mcI}=\set{[\cd,\ld,\cd]}$, that is, none of the elements is known. Then, \eqref{tilde.brho.M-1} places element $\hat{e}_1$ in the first position, which was not known in $\hat{\mcR}_{\mcI}$. There may be many preliminary rankings associated with a node $\mcI'$, so $\hat\mcR_{\mcI'}$ is, in general, a \emph{list} of preliminary rankings.

We gather all degree-compatible nodes $\mcI\backslash\hat{e}_1$ in the subfamilies $\mcA_{M-1}\subset\mcF_{M-1}$. Next, for each node $\mcI'\in\mcA_{M-1}$ we look for the degree-compatible parent nodes $\mcI'' = \mcI'\backslash \hat{e}_2$, all of which we gather in $\mcA_{M-2}$. Each of the parent nodes $\mcI''$ will ``inherit" the list of preliminary rankings from its children, and each preliminary rankings on such list will have $\hat{e}_2$ placed at the second position:
\begin{align}
\label{tilde.brho.tmp}
\hat\mcR_{\tr{tmp}}
&\leftarrow\hat\mcR_{\mcI'\backslash\hat{e}_2}\\
\label{tilde.brho.M-2}
 \rho_2 
 &\leftarrow \hat{e}_2, \quad \forall \brho \in \hat\mcR_{\tr{tmp}},\\
 \label{merge.subsets}
 \hat\mcR_{\mcI'\backslash\hat{e}_2}
 &\leftarrow
 \hat\mcR_{\mcI'\backslash\hat{e}_2}
 \cup
 \hat\mcR_{\tr{tmp}},
\end{align}
where \eqref{tilde.brho.tmp}-\eqref{merge.subsets} take care of the situation when a parent node has many degree-compatible children. If this happens, the parent node inherits all lists of preliminary rankings from its children, sets the corresponding element of each ranking on the list via \eqref{tilde.brho.M-2}, and then merges the modified lists via \eqref{merge.subsets}.

The recursion can be described by the pseudo-code shown in Algorithm~\ref{Algo:orders} available at \github.

% ------------------------------- algorithm -----------------------
\begin{algorithm}
    \caption{Finding all solutions $\hat\brho$} \label{Algo:orders}
\begin{algorithmic}[1]
\State Input: $\matW\in\Natural^{M\times M}$ \Comment Matrix of the results

\textbf{Forward recursion:}
\State Initialization:
\State$\mcF_1=\set{\{1\},\{2\},\ld, \{M\}}$\label{DEF.F.1.Sol} \Comment subsets of $\mcI$ with one element
\For{$\mcI'\in \mcF_1$}\label{FOR.I.prim.1.q}
    \State $q(\mcI')\leftarrow 0$
\EndFor
\State Main recursion loop:
\For{ $k = 2, \ld, M$ }\label{FOR.k.q}
    \State Create $\mcF_k$ \Comment all subsets of $\mcI$ with $k$ elements, $|\mcF_k|=\nchoosek{M}{k}$
    \For{$\mcI' \in\mcF_k$}\label{FOR.mcI.prim.q}
        % \State $q(\mcI')\leftarrow 0$
        % \For{$e\in\mcI'$}    \label{FOR.e}
	       \State $q(\mcI') \leftarrow  \max_{e\in\mcI'}\set{d_e(\mcI') +q(\mcI'\backslash e)}$ \label{LINE.polynomial.sol}
    \EndFor
\EndFor % i

\textbf{Backward recursion:}
\State Initialization: 
\State $\mcA_M \leftarrow \set{\mcI}$, 
\State $\hat\mcR_{\mcI}\leftarrow\set{[\cd,\ld,\cd]}$\Comment The ranking is unknown at this moment
\State $\mcI_{\tr{o}}\equiv\set{}$
\State $\mcF_{0}\equiv\set{\mcI_{\tr{o}}}$
\State Backward search
\For{ $k = M,\ld, 1$ }\label{FOR.k.q.2}
    \State $\mcA_{k-1}\leftarrow \set{}$ \Comment Subfamily of degree-compatible nodes $\mcI''\in\mcF_{k-1}$,
    % \For{$\mcI''\in\mcF_{k-1}$}
        % \State $\hat\mcR_{\mcI''}\leftarrow \set{}$
        % \Comment No preliminary solutions associated with 
            $\mcI''$
    % \EndFor
    \For{$\mcI' \in \mcA_k$}\label{FOR.mcI.prim.q.back}
        \For{$e\in\mcI'$}    \label{FOR.e.back}
            \State $\mcI'' \leftarrow \mcI' \backslash e$
            \If{$q(\mcI')- d_{e}(\mcI')=q(\mcI'') $}
                %\State $\hat{e}\leftarrow e$ 
                %\Comment $\mcI'$ and $\mcI''$ are degree-compatible
                \State $\hat\mcR_{\tr{tmp}} \leftarrow \hat\mcR_{\mcI'}$
                \For{$\brho\in \hat\mcR_{\tr{tmp}}$}
                    \State $\rho_k \leftarrow \hat{e}$
                    \Comment  Set $k$-th element to $e$ for all orders in $\hat\mcR_{\tr{tmp}}$ 
                \EndFor
                \If{$\mcI''\in \mcA_{k-1}$}
                    \State $\hat\mcR_{\mcI''} \leftarrow  \hat\mcR_{\mcI''}\cup \hat\mcR_{\tr{tmp}}$ \Comment Merge lists of preliminary solutions
                \Else
                    \State $\mcA_{k-1}\leftarrow \mcA_{k-1} \cup \mcI''$
                    \State $\hat\mcR_{\mcI''} \leftarrow  \hat\mcR_{\tr{tmp}}$ \Comment Create lists of preliminary solutions
                \EndIf
            \EndIf
        \EndFor
    \EndFor
\EndFor

\State Output: $\hat\mcR_{\mcI_{\tr{o}}}$   \Comment contains all the solutions $\hat\brho$ 
\end{algorithmic}
\end{algorithm}
% ------------------------------- end algorithm -------------------

%%%%%%%%%%%%%%
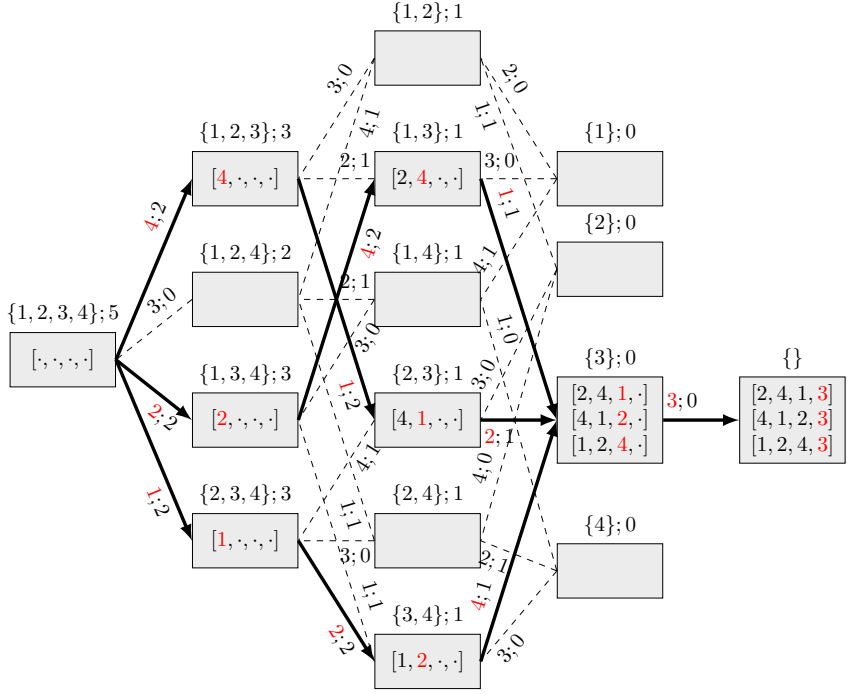
\begin{figure}[bt]
\begin{center}
\scalebox{\sizf}{%!TEX root =  ../Phase.Track.HARQ.tex
\pgfdeclarelayer{background}
\pgfdeclarelayer{foreground}
\pgfsetlayers{background,main,foreground}

\tikzstyle{rec_my3} = [draw, rectangle, minimum width=1.2cm, text width=1.5cm, fill=gray!15, text centered,  minimum height=.9cm]

\tikzstyle{line_my3} = [draw, latex-, ultra thick] 
\tikzstyle{line_my4} = [draw, dashed, line width=0.1mm] 
%=================
% \begin{center}
\begin{tikzpicture}[trim left=0cm,node distance=3cm]

  %%%%%--------  Set nodes -------- %%%%%%%%%%%%%%%%  
  % F_4
\node[rec_my3]	(I_4) 		[label=north:{$\set{1,2,3,4}; 5$}]	{$[\cd,\cd,\cd,\cd]$};
% F_3
\node[rec_my3] 	(I_3_1) 	[right of=I_4,node distance=3cm,yshift=3cm,label=north:{$\set{1,2,3}; 3$}]	{$[\red{4},\cd,\cd,\cd]$};
\node[rec_my3] 	(I_3_2) 	[below of=I_3_1,yshift=1cm,label=north:{$\set{1,2,4}; 2$}]	{};
\node[rec_my3] 	(I_3_3) 	[below of=I_3_2,yshift=1cm,label=north:{$\set{1,3,4}; 3$}]	{$[\red{2},\cd,\cd,\cd]$};
\node[rec_my3] 	(I_3_4) 	[below of=I_3_3,yshift=1cm,label=north:{$\set{2,3,4}; 3$}]	{$[\red{1},\cd,\cd,\cd]$};
% F_2
\node[rec_my3] 	(I_2_1) 	[right of=I_3_1,node distance=3cm, yshift=2cm,label=north:{$\set{1,2}; 1$}]	{};
\node[rec_my3] 	(I_2_2) 	[below of=I_2_1,yshift=1cm,label=north:{$\set{1,3}; 1$}]	{$[2,\red{4},\cd,\cd]$};
\node[rec_my3] 	(I_2_3) 	[below of=I_2_2,yshift=1cm,label=north:{$\set{1,4}; 1$}]	{};
\node[rec_my3] 	(I_2_4) 	[below of=I_2_3,yshift=1cm,label=north:{$\set{2,3}; 1$}]	{$[4,\red{1},\cd,\cd]$};
\node[rec_my3] 	(I_2_5) 	[below of=I_2_4,yshift=1cm,label=north:{$\set{2,4}; 1$}]	{};
\node[rec_my3] 	(I_2_6) 	[below of=I_2_5,yshift=1cm,label=north:{$\set{3,4}; 1$}]	{$[1,\red{2},\cd,\cd]$};
% F_1
\node[rec_my3] 	(I_1_1) 	[right of=I_3_1,node distance=6cm, label=north:{$\set{1}; 0$}]	{};
\node[rec_my3] 	(I_1_2) 	[below of=I_1_1,yshift=1.5cm,label=north:{$\set{2}; 0$}]	{};
\node[rec_my3] 	(I_1_3) 	[below of=I_1_2,yshift=0.5cm,label=north:{$\set{3}; 0$}]	{$[2,4,\red{1},\cd]$\\$[4,1,\red{2},\cd]$\\$[1,2,\red{4},\cd]$};
\node[rec_my3] 	(I_1_4) 	[below of=I_1_3,yshift=0.5cm,label=north:{$\set{4}; 0$}]	{};
% F_0
\node[rec_my3] 	(I_0_1) 	[right of=I_1_3, label=north:{$\set{}$}]	{$[2,4,1,\red{3}]$\\$[4,1,2,\red{3}]$\\$[1,2,4,\red{3}]$};

 %%%%%--------    Draw Paths  -------- %%%%%%%%%%%%%%%%
 % F_0 to F_1
\path[line_my3]   (I_0_1.west)	-- node[above, near end, sloped]{$\red{3};0$}	(I_1_3.east);
 % F_1 to F_2
\path[line_my4]   (I_1_1.west)	-- node[above, near end, sloped]{$2;0$}	(I_2_1.east);
\path[line_my4]   (I_1_1.west)	-- node[above, near end, sloped]{$3;0$}	(I_2_2.east);
\path[line_my4]   (I_1_1.west)	-- node[above, near end, sloped]{$4;1$}	(I_2_3.east);
\path[line_my4]   (I_1_2.west)	-- node[below, near end, sloped]{$1;1$}	(I_2_1.east);
\path[line_my4]   (I_1_2.west)	-- node[above, near end, sloped]{$3;0$}	(I_2_4.east);
\path[line_my4]   (I_1_2.west)	-- node[above, near end, sloped]{$4;0$}	(I_2_5.east);
\path[line_my3]   (I_1_3.west)	-- node[above, very near end, sloped]{$\red{1};1$}	(I_2_2.east);
\path[line_my3]   (I_1_3.west)	-- node[below, near end, sloped]{$\red{2};1$}	(I_2_4.east);
\path[line_my3]   (I_1_3.west)	-- node[above, near end, sloped]{$\red{4};1$}	(I_2_6.east);
\path[line_my4]   (I_1_4.west)	-- node[above, very near end, sloped]{$1;0$}	(I_2_3.east);
\path[line_my4]   (I_1_4.west)	-- node[below, near end, sloped]{$2;1$}	(I_2_5.east);
\path[line_my4]   (I_1_4.west)	-- node[below, near end, sloped]{$3;0$}	(I_2_6.east);
% F_2 to F_3
\path[line_my4]   (I_2_1.west)	-- node[above, near start, sloped]{$3;0$}	(I_3_1.east);
\path[line_my4]   (I_2_2.west)	-- node[above, near start, sloped]{$2;1$}	(I_3_1.east);
\path[line_my3]   (I_2_4.west)	-- node[below, very near start, sloped]{$\red{1};2$} (I_3_1.east);
\path[line_my4]   (I_2_1.west)	-- node[below, near start, sloped]{$4;1$}	(I_3_2.east);
\path[line_my4]   (I_2_3.west)	-- node[above, near start, sloped]{$2;1$}	(I_3_2.east);
\path[line_my4]   (I_2_5.west)	-- node[below, very near start, sloped]{$1;1$} (I_3_2.east);
\path[line_my3]   (I_2_2.west)	-- node[below, near start, sloped]{$\red{4};2$}	(I_3_3.east);
\path[line_my4]   (I_2_3.west)	-- node[below, near start, sloped]{$3;0$}	(I_3_3.east);
\path[line_my4]   (I_2_6.west)	-- node[above, near start, sloped]{$1;1$} (I_3_3.east);
\path[line_my4]   (I_2_4.west)	-- node[below, near start, sloped]{$4;1$}	(I_3_4.east);
\path[line_my4]   (I_2_5.west)	-- node[below, near start, sloped]{$3;0$}	(I_3_4.east);
\path[line_my3]   (I_2_6.west)	-- node[below, near start, sloped]{$\red{2};2$} (I_3_4.east);
% F_3 to F_4
\path[line_my3]   (I_3_1.west)	-- node[above, near start, sloped]{$\red{4};2$}	(I_4.east);
\path[line_my4]   (I_3_2.west)	-- node[above, near start, sloped]{$3;0$}	(I_4.east);
\path[line_my3]   (I_3_3.west)	-- node[below, near start, sloped]{$\red{2};2$}	(I_4.east);
\path[line_my3]   (I_3_4.west)	-- node[below, near start, sloped]{$\red{1};2$}	(I_4.east);

%%%% %% --- Draw Big Box
%\begin{pgfonlayer}{background}
%\node[bigbox_my,fill=blue!20, fit=(Dem1)(Demk)(Agg),label=north:$\DEMwo$]   (Dem_tot) {};
%\node[bigbox_my,fill=blue!20, fit=(Chan1)(Chank),label=north:Channel]   (Chan_tot) {};
%\node[below right] at (RX.north)  {Receiver};
%\end{pgfonlayer}

%%%%
\end{tikzpicture}
% \end{center}
}
\end{center}
\caption{Backward recursion of Algorithm~\ref{Algo:orders} applied to the matrix $\matW$ shown in \exref{Example:Slater.spectrum}. We show $\mcI',q(\mcI')$ above the node. The vectors inside the nodes represent the (list of) preliminary rankings $\hat\mcR_{\mcI'}$ which are created in the backward recursion. The thick edges connect degree-compatible nodes while the dashed ones connect the nodes that are not visited in the backward recursion (so these nodes are empty). The edges leaving $\mcI'$ rightward to $\mcI''=\mcI'\backslash e\in\mcF_{k-1}$ are labeled as ``$e;d_e(\mcI')$", where $e$ is shown in red on the edges connecting the degree-compatible nodes, and it becomes the $k$-th elements of the preliminary ranking in $\hat\mcR_{\mcI''}$. The unique, right-most node (indexed with an empty set $\set{}$) contains all optimal rankings $\hat\brho$.}
\label{fig:Algorithm.backward}
\end{figure}
%%%%%%%%%%%%%%%

\textbf{Complexity}
Note that the entire algorithm we describe does not make any reference to the spectrum. It only uses the orders of the polynomials and is truly an aside result obtained due to the particular formulation of the problem. 

Removing the calculation of the spectra, it is quite easy to see that the complexity may be evaluated as 
\begin{align}
    \tr{CMPLX} 
    &= \sum_{k=2}^M \nchoosek{M}{k} k\\%= M(M-1)2^{M-2}\\
    &=O\Big(M 2^M \Big),%O\Big( (M^2 + K)2^M \Big),
\end{align}
and the complexity of the backward recursion in which we recover the solutions is neglected because we only must visit order-compatible nodes from the subfamilies $\mcA_k$, and these are significantly smaller than $\mcF_k$.

Using the method shown in \cite[Sec.~3]{Remage66}, the complexity of finding only \emph{one} solutions $\hat\brho$ is $O(M 2^{M})$. Therefore, it is interesting to see that it is possible to find \emph{all} the solutions with the same cost. Of course, we ignore all the cost related to indexing, as also done in \cite{Remage66}.

\begin{example}[Recovering all rankings]\label{Ex:Recovering.solutions}

We use the same matrix $\matW$ as in \exref{Example:Slater.spectrum} and illustrate both the forward and backward recursion steps in \figref{fig:Algorithm.backward}. 

In the forward recursion, we are only interested in the calculation of the degrees of the polynomials $q(\mcI')$ that are easily obtained applying \eqref{recursive.order} and are shown above the nodes together with the sets as ``$\mcI; q(\mcI')$". Although this is the most computationally intensive part of the algorithm (because we have to visit all nodes), it is rather straightforward. 

The backward phase is more interesting and starts with the node $\mcI$ whose preliminary ranking is entirely unknown, \ie $\hat{\mcR}_{\mcI}=\set{[\cd,\ld,\cd]}$ as shown inside the node.

We then identify the degree-compatible nodes that satisfy \eqref{order.index} and form the subfamily $\mcA_{3}=\set{\set{1,2,3}, \set{1,3,4}, \set{2,3,4}}$. The first element of the label ``$e; d_e{\mcI'}$" on the edge (the thick arrow connecting the degree-compatible nodes $\mcI'$ and $\mcI''=\mcI'\backslash{e}$) is shown in red, and it fills the left-most (\ie the first) undetermined position in the preliminary ranking $\hat{R}_{\mcI''}$.

The same step is repeated to find the subfamily $\mcA_{2}$, where the preliminary rankings associated with the children nodes $\mcI\in\mcA_{3}$ are inherited by the parents in the subfamily $\mcA_2$. Since the first position of the ranking is already known, the edge label is used to determine the second element in the preliminary ranking.

The more interesting situation arises when we find the subfamily $\mcA_1=\set{\set{3}}$ which has only one element, the set $\mcI''=\set{3}$ which has three children in the subfamily $\mcA_2$, from whom it inherits three preliminary rankings and uses the labels on the edges to determine the third position in each of them.

In the last step, we gather all the preliminary rankings and fill the last missing position with the element on the edge. In our example, there is only one such element, $e=3$.

Again, the steps shown in \figref{fig:Algorithm.backward} can be reproduced using the program shown in \github.

\end{example}

%%%%%%%%%%%%%%%%%%%%%%%%
% \subsection{Recursive calculation of the rankability criterion $\sigma$}\label{Sec:recursive.sigma}

%%%%%%%%%%%%%%%%%%%%%%%%%%%%%%%%%%%%%%%%%%%%%%%%%%%%%%%
\section{Numerical examples}\label{Sec:Numerical.Examples}
%%%%%%%%%%%%%%%%%%%%%%%%%%%%%%%%%%%%%%%%%%%%%%%%%%%%%%%
\subsection{Synthetic data}\label{Sec:Synthetic}
To better understand the estimation problem at hand, we randomly generate the data with the value of the parameter $p=\ov{p}$, where we control $\ov{p}$. We use $M=10$ and $K_{m,n}=2$ so that $T=M(M-1)$, which is similar to the sports league scenario in which teams play against each other twice (typically once at home and once as visitors). 

\begin{figure}[ht!]
\centering
\includegraphics[width=0.8\linewidth]{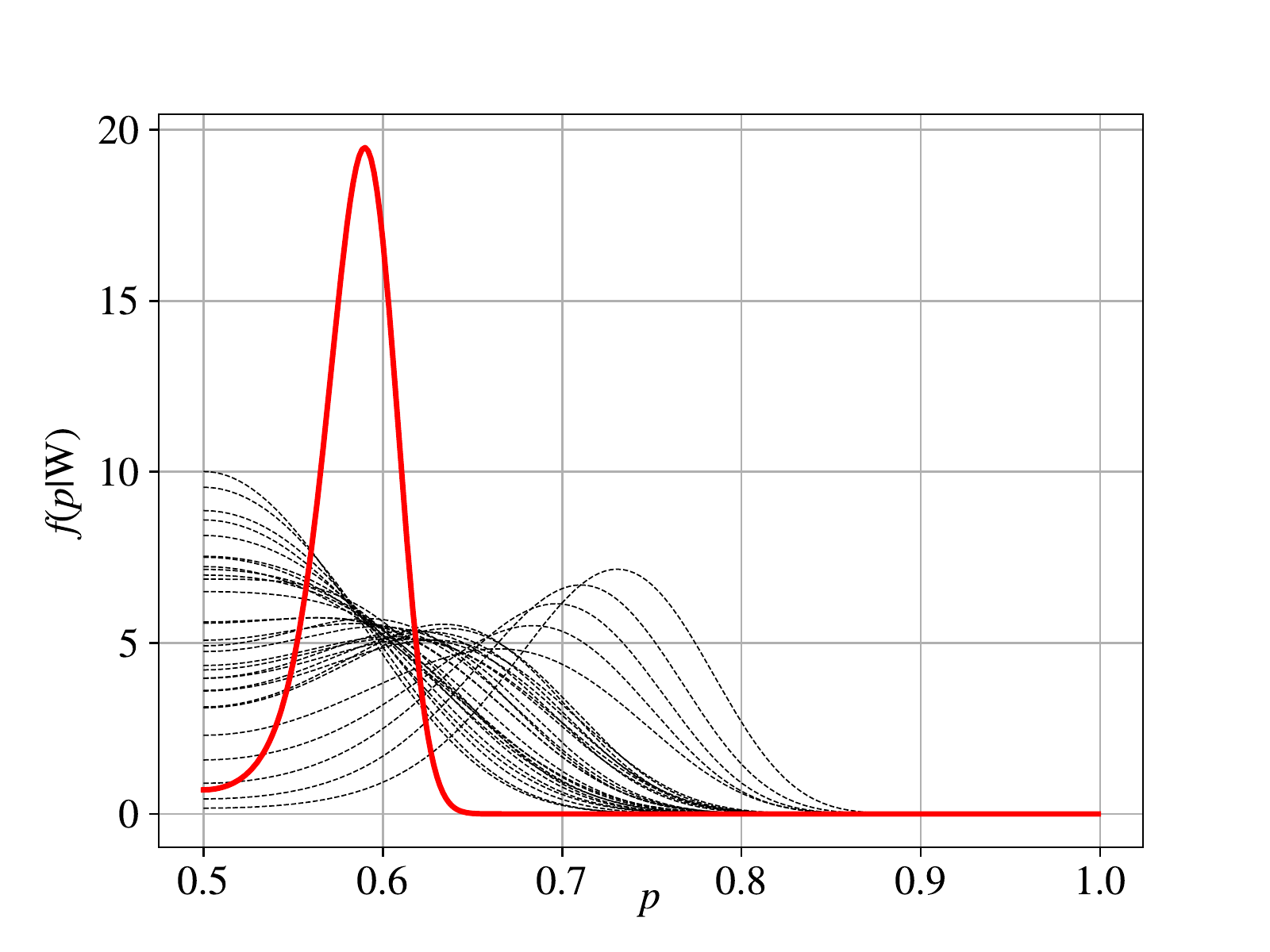}\\
a)\\

\includegraphics[width=0.8\linewidth]{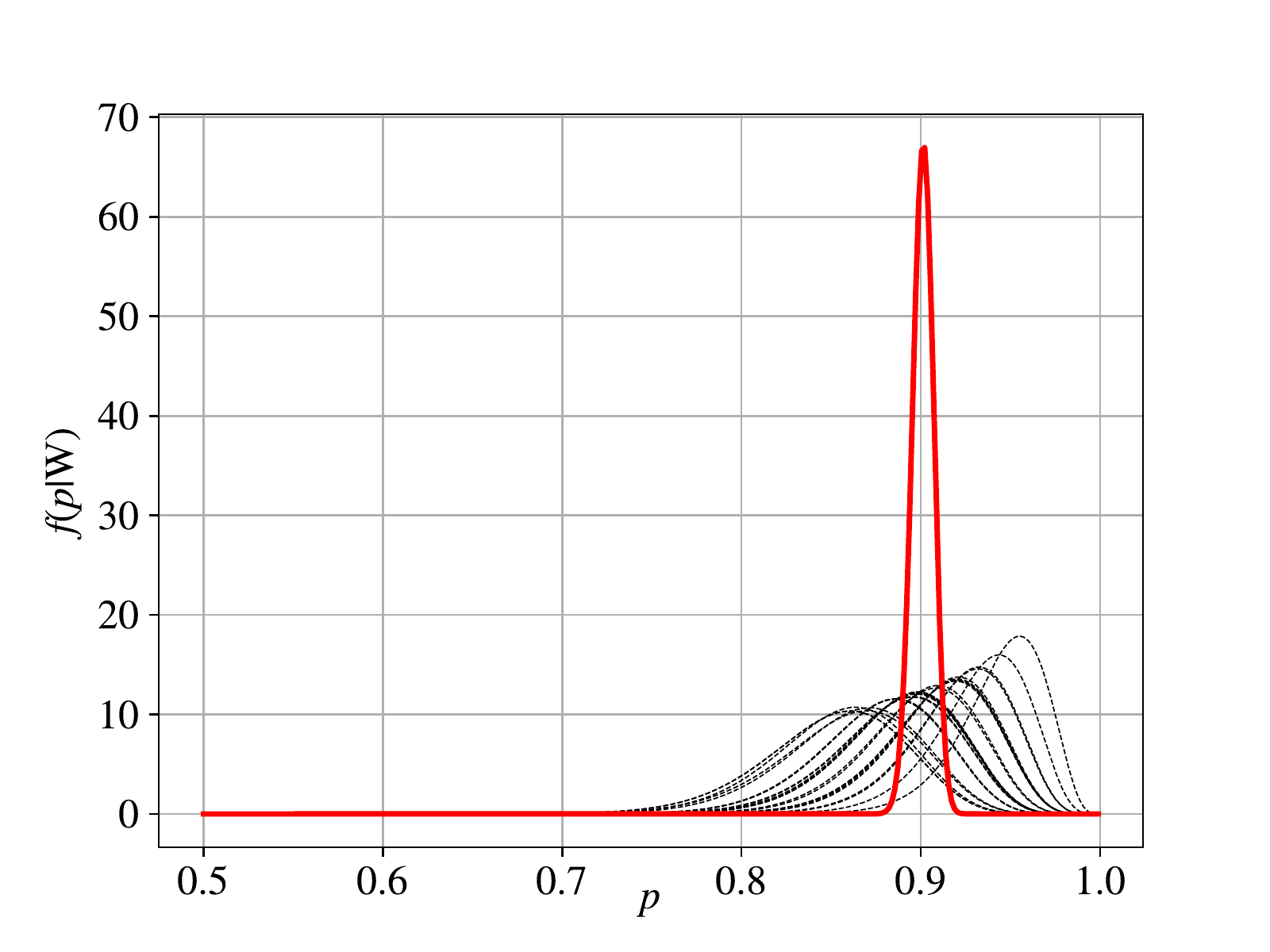}\\
b)
\caption{Posterior \gls{pdf}s $\pdf(p|\matW_l), l=1,\ld,L$, $l=30$, obtained for randomly generated matrices $\matW_l$; ($M=10$, $K_{m,n}=2$) with nominal value a) $\ov{p}=0.6$ and b) $\ov{p}=0.9$. The thick red line indicates the \gls{pdf} $\pdf(p|\set{\matW_l}_{l=1}^L)$. }\label{Fig:pdf_0.6}
\end{figure}

\figref{Fig:pdf_0.6} shows the posterior distributions $\pdf(p|\matW_l)$ for $L=30$ random realizations of the matrix $\matW_l$. These are, of course, only examples, because the function $\pdf(p|\matW)$, depending on the randomly generated $\matW$, is also random. The \gls{pdf} $\pdf(p|\set{\matW_l}_{l=1}^L)$, \cf \eqref{product.phi}, is shown with a thick red line.

We observe that the information about $p$ obtained from a particular realization $\matW_l$ may be very unreliably, which is particularly visible for small nominal $\ov{p}$, \eg for $\ov{p}=0.6$, the mode varies significantly, $\check{p}_l\in[0.5,0.75)$. In particular, many different realizations $\matW_l$ produce the \gls{pdf} with the mode $\check{p}=0.5$; this means that the most likely explanation is that these matrices are not rankable. The variability is much smaller for larger $\ov{p}$, \eg $\ov{p}=0.9$. This is also visible in the form of the \gls{pdf} estimated from all realizations, $\pdf(p|\set{\matW_l}_{l=1}^L)$, where, although we obtain the distributions centered on $\ov{p}$, the variance (assessed by the width of the distribution) is also larger for smaller $\ov{p}$.

\begin{figure}[bt!]
\centering
\includegraphics[width=0.8\linewidth]{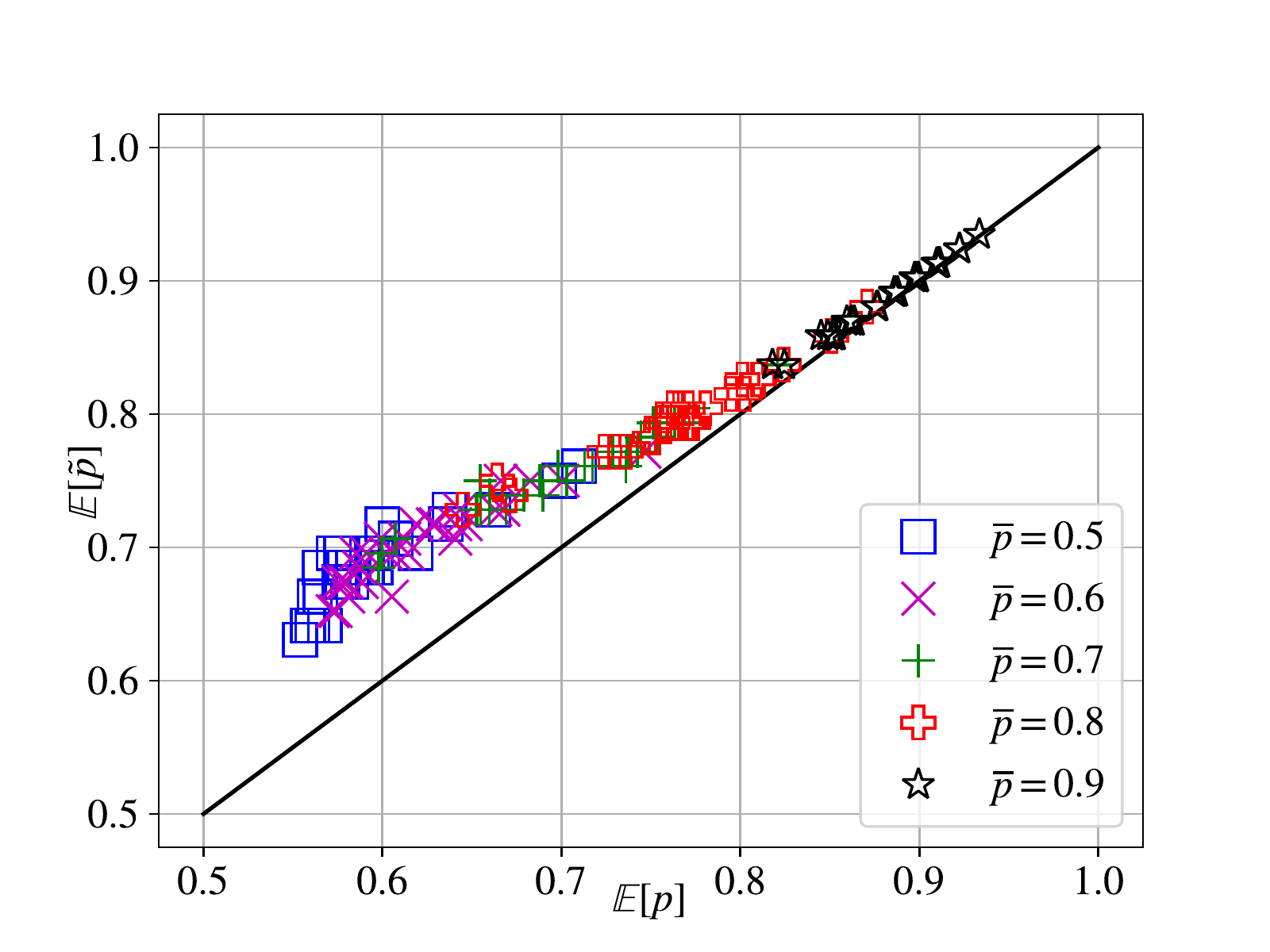}\\
a)\\
\includegraphics[width=0.8\linewidth]{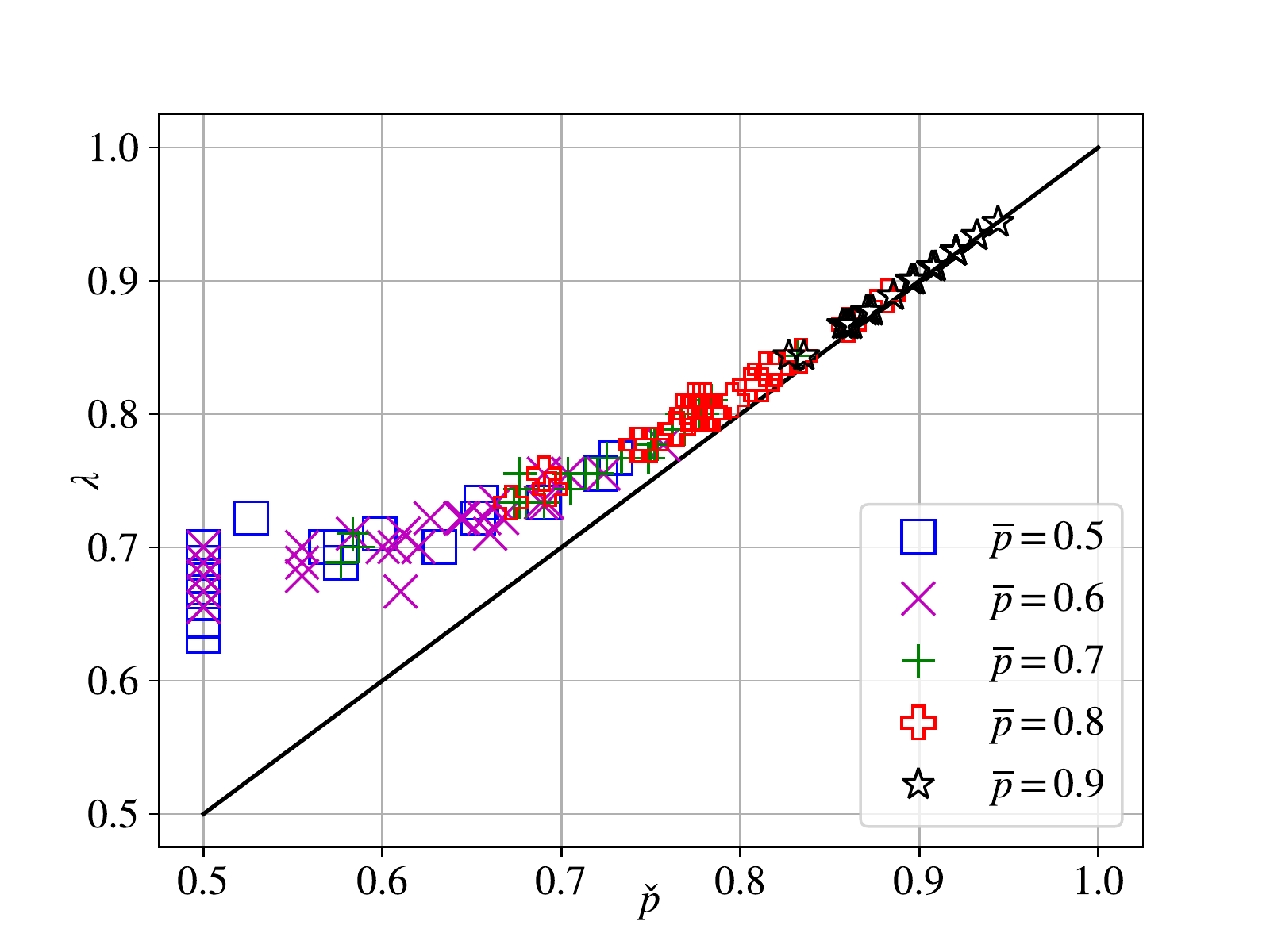}\\
b)
\caption{Relationship between the parameters of the  \gls{pdf}s $\pdf(p|\matW_l)$ and $\pdf(\tilde{p}|\matW_l)$ calculated for randomly generated matrices $\matW_l, l=1,\ld,L$  with the nominal value $\ov{p}\in\set{0.5,0.6,0.7,0.8, 0.9}$ a) $\Ex[\tilde{p}]$ vs $\Ex[p]$, and b) $\lambda$ vs $\check{p}$. }\label{Fig:mean_mode}
\end{figure}

To illustrate the relationship between the parameters of the \gls{pdf} $\pdf(p|\matW_l)$ and its approximation $\pdf(\tilde{p}|\matW_l)$, the relationships between the posterior means and posterior modes are shown, respectively, in \figref{Fig:mean_mode}a and \figref{Fig:mean_mode}b. 

These results,  obtained for $L=30$ realizations of $\matW_l$ with different values of $\ov{p}\in\set{0.5,0.6,0.7,0.8,0.9}$, confirm our analysis from \secref{Sec:Posterior.Mean} and \secref{Sec:Mode}: for \gls{pdf}s concentrated in the vicinity of $p=1$ (these cases are obtained when $\ov{p}=0.9$, see \ref{Fig:pdf_0.6}b), the approximation of the \gls{pdf} with the unique term corresponding to the Slater index $\hat{S}$ is accurate. Then, both the mean and the mode are accurately estimated, and, are close to each other; this is not the case for non-rankable data. Consequently, the degree of linearity $\lambda$ is then a suitable indicator of rankability.

On the other hand, using only the Slater index to detect non-rankable $\matW$ seems difficult when $\ov{p}$ is small. Indeed, in \figref{Fig:mean_mode}b, all cases where $\check{p}=0.5$ yield $\lambda\in(0.62,0.7)$ while the approximate threshold \eqref{lambda.threshold}, which allows us to assume that $\check{p}>0.5$ is given by $\lambda_{\tr{th}}=0.55$.

It is also worthwhile to note the significant overlap between the solutions obtained for $\ov{p}=0.5$ and $\ov{p}=0.6$, which begs a question: should rankability be determined from a particular realization of $\matW$, or rather, concluded, when possible, from multiple realizations of $\matW_l$? We do not have an answer to that conundrum.

Clearly, the issue of finding the suitable threshold to declare the data as non-rankable requires more research but the pragmatic, although conservative (avoiding false positives) solution is to consider data rankable only for sufficiently large $\lambda$, \eg $\lambda>0.8$ in the examples we have shown.

\begin{figure}[hbt!]
\centering
\includegraphics[width=0.8\linewidth]{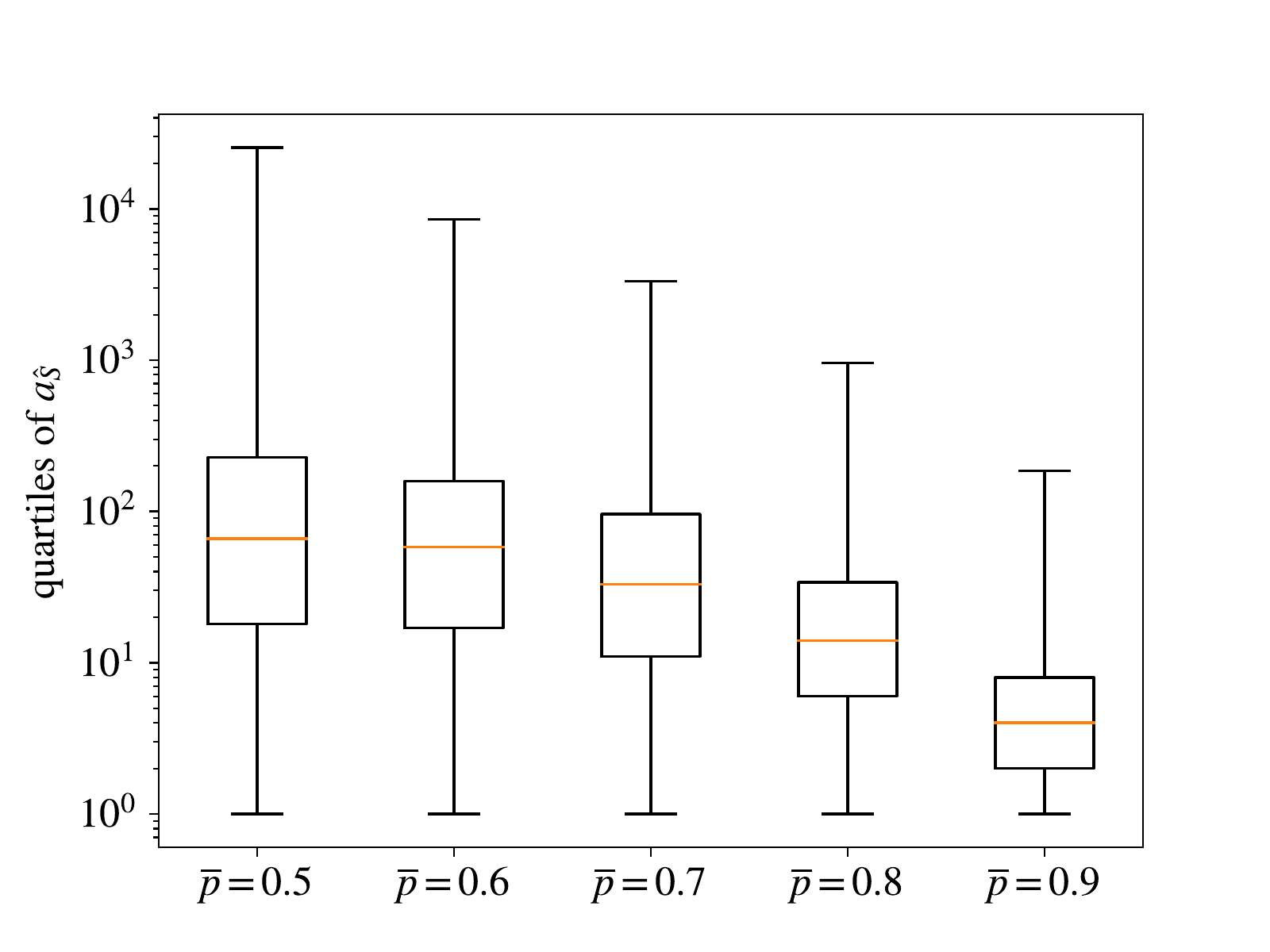}\\
\caption{Quartiles of the empirical distribution of $a_{\hat{S}}$ for  $\ov{p}\in\set{0.5,0.6,0.7,0.8,0.9}$; whiskers indicate the range of the observed values. }\label{Fig:a_S}
\end{figure}

We terminate the exploration of the synthetic data by investigating the multiplicity of the solutions, \ie the value of the coefficient $a_{\hat{S}}$. To this end, we reproduce the experiments we run for $L=1000$ and show the quartiles (as a box-plot) in \figref{Fig:a_S}; note the logarithmic scale.

We can easily see that the probability of having a large multiplicity $a_{\hat{S}}$ decreases with $\ov{p}$. By correlating this observation with \figref{Fig:mean_mode} we may conclude that the intuition of \cite{Anderson19} is correct: not only $\hat{S}$, but also $a_{\hat{S}}$ are evidence of rankability. This empirical observation clashes with our discussion about the degenerate spectrum (see text after \eqref{tilde.Z}) which concluded that $a_{\hat{S}}$ is irrelevant. Therefore, it is possible that there are other more informative forms of the Slater spectrum that can be inferred from the knowledge of the pair $(\hat{S}, a_{\hat{S}})$. We must leave this question as an open problem.

% \begin{figure}[bt]
% \centering
% \begin{tabular}{cc}
% \includegraphics[width=0.45\linewidth]{./figures/phi_func_K=2_p=0.6_0.9.pdf}
% &
% \includegraphics[width=0.45\linewidth]{./figures/phi_func_K=2_p=0.5.pdf}\\
% a) $\ov{p}\in\set{0.6,0.9}$ & b) $\ov{p}=0.5$
% \end{tabular}
% \caption{Posterior \gls{pdf}s $\pdf(p|\matW)$ obtained for randomly generated matrices $\matW$; $M=10$, and (a) $\ov{p}=0.6$ (solid lines) or $\ov{p}=0.9$ (dashed lines), and b) $\ov{p}=0.5$.}\label{Fig:phi_func}
% \end{figure}

% We observe in \figref{Fig:phi_func}a that even if the data are generated with $\ov{p}=0.6$, it is still possible to obtain the estimate $\check{p}=0.5$ (the mode of the \gls{pdf}). Similarly, by generating data with $\ov{p}=0.5$ whose results are plotted in \figref{Fig:phi_func}b, we see that despite the fact that the results are entirely random (and therefore not related to the order), we can still obtain estimates $\check{p}>0.5$. 

%%%%%%%%%%%%%%%%%%%%%%%%%%%%%%%%%%%%%%%%%%%%%%%%%%%%%%%
\subsection{Real-world data}\label{Sec:Real-world}
We will now use real-world data obtained from sports competitions with binary outcomes (wins / losses) and consider ten seasons of cricket games in the \gls{ipl} and ten seasons in the Italian professional volleyball SuperLega. The summary of the data and the results obtained are shown in \tabref{Tab:summary}. In principle, there are $K=2$ games per pair of teams and thus $T=M(M-1)$ but, in some seasons of \gls{ipl}, the games were abandoned and therefore $T$ is different as we can see in \tabref{Tab:summary}a. 

\begin{figure}[ht!]
\centering
\includegraphics[width=0.8\linewidth]{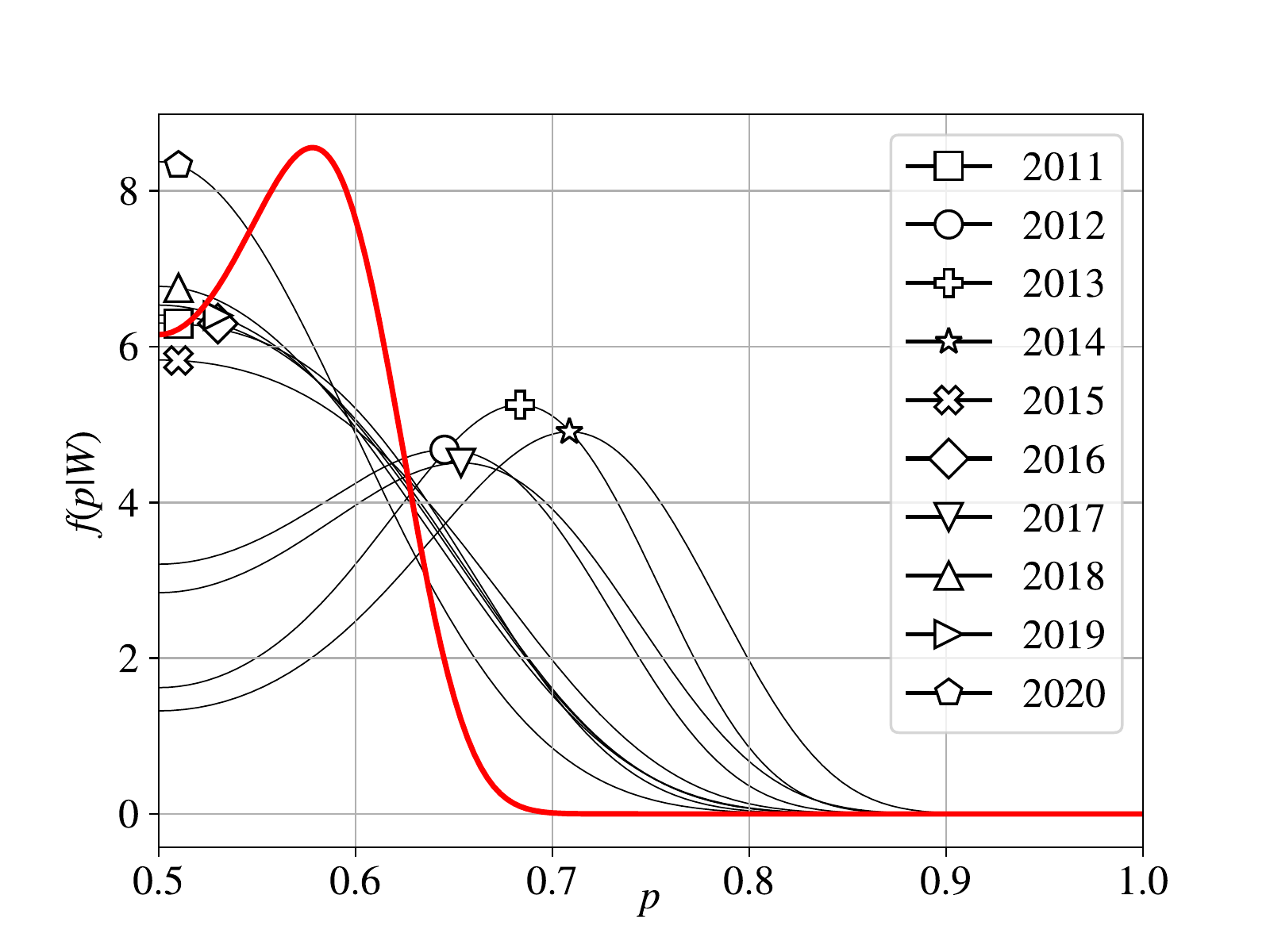}\\
a) \gls{ipl} \\

\includegraphics[width=0.8\linewidth]{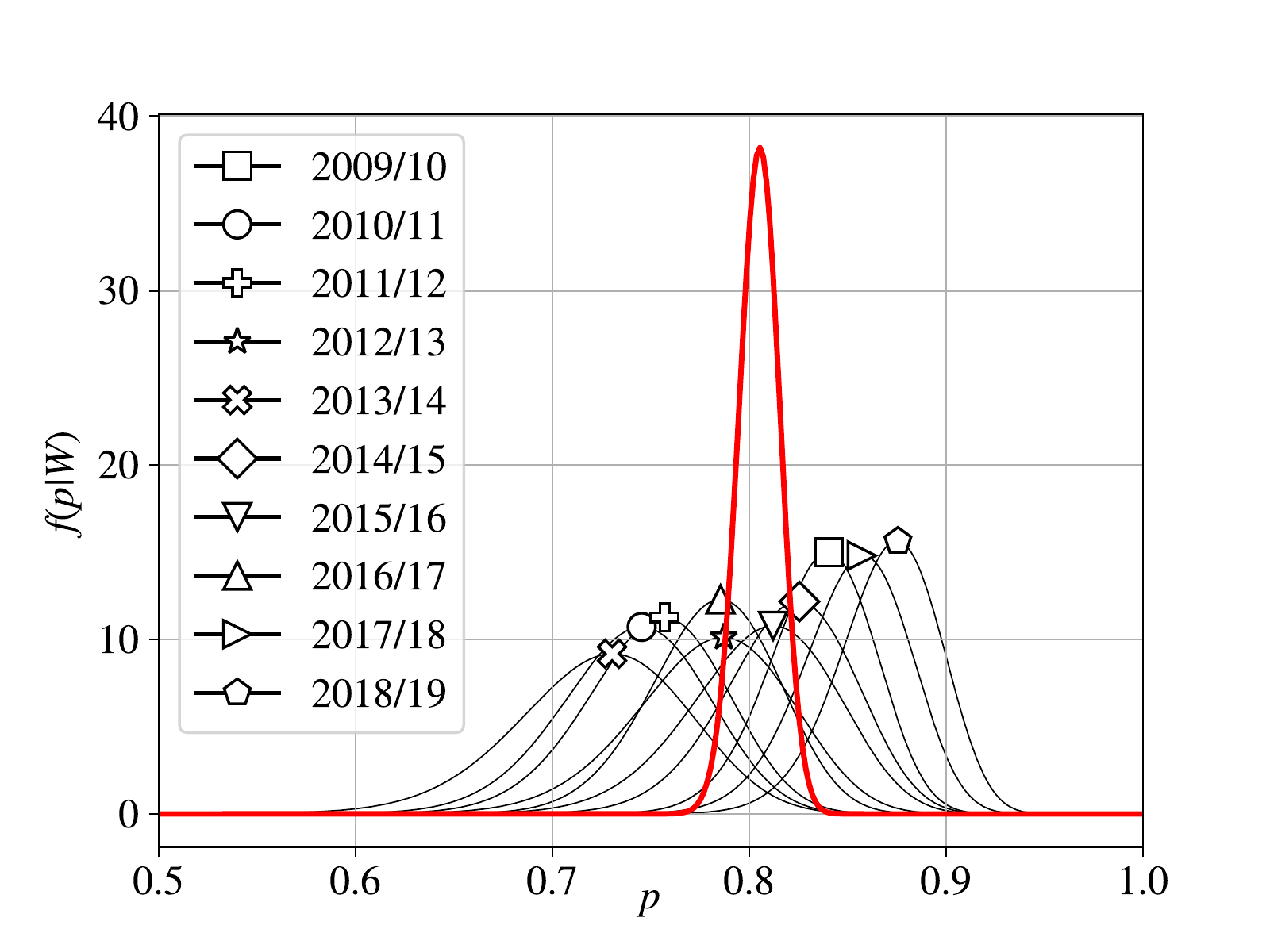}\\
b) SuperLega
\caption{Posterior \gls{pdf} $\pdf(p|\matW_l), l=1,\ld,L$ obtained from results of $L=10$ seasons of a) \gls{ipl} cricket and b) Italian SuperLega volleyball. The thick red line indicates the \gls{pdf} $\pdf(p|\set{\matW_l}_{l=1}^L)$.}\label{Fig:phi_func_real}
\end{figure}

The \gls{pdf}s obtained for the \gls{ipl} data are shown in \figref{Fig:phi_func_real}a and those for SuperLega data in \figref{Fig:phi_func_real}b the \gls{pdf} obtained jointly from all seasons is shown with a thick red line.

\begin{table}[ht!]
    \caption{Data and results in \gls{ipl} and SuperLega; note that, in the former, we may have different values of $T$ for the same $M$ (due to canceled games). The Slater index from \eqref{ML.definition.final.C.hat}, $\hat{S}$ (shown) is used to calculate the degree of linearity $\lambda$ (shown) using \eqref{degree.linearity} which is an approximation of the mode $\check{p}$ (also shown) defined in \eqref{p.MAP}. The number of optimal rankings $\hat\brho$ \eqref{ML.definition.final.C} is given by the first non-zero element of the Slater spectrum $a_{\hat{S}}$ (shown). The line labeled as ``joint" shows only the mode $\check{p}$ of $\pdf(p|\set{\matW_l}_{l=1}^L)$ and the joint degree of linearity $\lambda_{\tr{joint}}$ obtained via \eqref{degree.linearity.joint}.
    \label{Tab:summary}
    }
    \centering
%%%%%%
\begin{tabular}{c|c|c|c|c|c|c}
season & $M$ & $T$ &  $\hat{S}$ & $a_{\hat{S}}$ & $\check{p}$ & $\lambda$/$\lambda_{\tr{joint}}$\\
    \hline
2011 & 10 & 68 & 20 & 15 & 0.50 & 0.71\\
2012 & 9 & 70 & 19 & 6 & 0.65 & 0.73\\
2013 & 9 & 72 & 19 & 52 & 0.68 & 0.74\\
2014 & 8 & 56 & 14 & 38 & 0.71 & 0.75\\
2015 & 8 & 53 & 16 & 21 & 0.50 & 0.70\\
2016 & 8 & 56 & 17 & 12 & 0.50 & 0.70\\
2017 & 8 & 55 & 15 & 33 & 0.65 & 0.73\\
2018 & 8 & 56 & 17 & 14 & 0.50 & 0.70\\
2019 & 8 & 55 & 17 & 14 & 0.50 & 0.69\\
2020 & 8 & 56 & 18 & 10 & 0.50 & 0.68\\
\hline
joint & $-$  & $-$  & $-$  & $-$  & 0.58 & 0.71
\end{tabular}
\\
a) \gls{ipl} \\

~

%%%%%
\begin{tabular}{c|c|c|c|c|c|c}
season & $M$ & $T$ &  $\hat{S}$ & $a_{\hat{S}}$ & $\check{p}$ & $\lambda$/$\lambda_{\tr{joint}}$\\
    \hline
2009/10 & 15 & 210 & 32 & 102 & 0.84 & 0.85\\
2010/11 & 14 & 182 & 40 & 5 & 0.75 & 0.78\\
2011/12 & 14 & 182 & 41 & 564 & 0.76 & 0.77\\
2012/13 & 12 & 132 & 26 & 86 & 0.79 & 0.80\\
2013/14 & 12 & 132 & 32 & 420 & 0.73 & 0.76\\
2014/15 & 13 & 156 & 25 & 20 & 0.83 & 0.84\\
2015/16 & 12 & 132 & 23 & 51 & 0.81 & 0.83\\
2016/17 & 14 & 182 & 37 & 3584 & 0.79 & 0.80\\
2017/18 & 14 & 182 & 25 & 93 & 0.86 & 0.86\\
2018/19 & 14 & 182 & 22 & 8 & 0.88 & 0.88\\
\hline
joint & $-$ & $-$ & $-$ & $-$  & 0.81 & 0.82
\end{tabular}
\\
b) SuperLega
\end{table}

% We show an example of the Slater spectrum in \figref{Fig:spectrum} obtained in 2016 season of Superlega where the symmetric form of the spectrum may be well appreciated and the non-zero coefficients of the spectrum are found only for $\hat{S}\le t \le T-\hat{S}$.

% \begin{figure}
%     \centering
%     \includegraphics[width=0.55\linewidth]{./figures/spectrum_SuperLega.pdf}
%     \caption{Example of the Slater spectrum $\set{a_t}$ obtained in 2016 season of Italian Superlega.}
%     \label{Fig:spectrum}
% \end{figure}

In the \gls{ipl} seasons 2011, 2015, 2016, 2018, 2019, and 2020, we obtain the \gls{ml} estimate $\check{p}=0.5$ (this is the mode of the \gls{pdf}), that is, the most likely explanation for the data is that the results are generated independently of any order. On the one hand, this is an interesting conclusion, which means that the league is well ``balanced'', and thus the games cannot be predicted and are interesting to watch. On the other hand, from the ranking perspective, this means that the solutions of the \gls{lop} cannot be meaningfully interpreted.

These observations may be contrasted with those obtained in the remaining seasons where $\check{p}>0.5$ and more insight may be obtained by observing jointly all seasons. Indeed, from both \figref{Fig:phi_func_real}a and \tabref{Tab:summary}a we can conclude that, since the joint \gls{pdf} is maximized for $\check{p}=0.58$. Indeed, these results are somewhat similar to those obtained from the synthetic data with $\ov{p}=0.6$. The behavior of the degree of linearity $\lambda$ is also similar: we see that it is much larger than the mode of the \gls{pdf} $\check{p}$.

The results obtained in the SuperLega are quite different and consistently show the estimate $\check{p}>0.7$ and the one obtained jointly, from all seasons, is $\check{p}=0.81$. This may be treated as confirmation that the ranking makes sense under the model \eqref{Pr=p}-\eqref{Pr=q}. 

Again, we may note the similarity to the synthetic data for $\ov{p}\in\set{0.8,0.9}$, where $\lambda$ closely follows the mode $\check{p}$.

The Slater index $\hat{S}$ is shown in \tabref{Tab:summary} together with the multiplicity of the equivalent optimal rankings, $a_{\hat{S}}$. Note that, while the former (or, equivalently, its transformation $\lambda$, see \eqref{degree.linearity}) is a relatively good indicator of the posterior mode (\eg $\lambda\le 0.71 \iff \check{p}=0.5$), this is not the case for the latter, which does not correlate well with $\check{p}$.

A similar conclusion is drawn using the data from SuperLega, where $a_{\hat{S}}$ varies significantly between seasons. For example, in the 2016/17 season we find the degree of linearity $\lambda=0.8$ (or $\hat{S}=37$) with $a_{\hat{S}}=3584$, while in the 2010/11 season, $\lambda=0.75$ with very small multiplicity $a_{\hat{S}}=5$. This large variability of $a_{\hat{S}}$ may preclude its reliable use for the detection of rankability. In fact, despite the simulation results shown in \figref{Fig:a_S}, which suggest that $a_{\hat{S}}$ may provide evidence for rankability, the above empirical observations are rather consistent with the analysis in \secref{Sec:Approximations.Slater.spectrum}, where we concluded that the multiplicity $a_{\hat{S}}$ does not affect the (degenerate) posterior \gls{pdf} (see text after \eqref{approximate.p=1.b}). This issue requires more study, as we are not able to resolve these apparent contradictions at this moment.

Finally, we also note that even if we can safely declare rankability in the case of SuperLega data, it is not immediately clear how multiple solutions should be interpreted. Indeed, we can easily recover all multiple rankings $\hat{\brho}$ using the algorithm shown in \secref{Sec:All.Solutions}, but it is not immediately obvious how to use them to rank the teams. Note that venues to solve this issue are proposed and studied in \cite{Anderson22}, however, our goal is not to determine whether or how the ranking should be done, but rather to point out that the results provided by our algorithms open the discussion on the properties of the data.

%%%%%%%%%%%%%%%%%%%%%%%%%%%%%%%%%%%%%%%%%%%%%%%%%%%%%%%
\section{Conclusions}\label{Sec:Conclusions}
In this work, we addressed the issue of defining the rankability of the data where, essentially, the goal is to make a statement about the reliability of solutions in the \acrfull{lop}. 

Our approach is probabilistic: we model pairwise comparison observations as Bernoulli variables characterized by a common parameter $p$ and find its estimate, and more generally, we obtain its posterior distribution. To this end, we defined the so-called Slater spectrum, which tells us how many orders share a given value of the consistency index; this is done for all possible values of the index. By construction, the Slater spectrum is a sufficient statistic for $p$, while the Slater index, often used in the literature, is not. An efficient algorithm is developed to find the Slater spectrum, which may be easily used to calculate the posterior distribution of the parameter $p$, which, in turn, tells us whether the \gls{lop} solution can be treated as a meaningful answer to the ordering problem. 

To simplify the analysis, we deal with the posterior mean and mode, and it allows us to link to the Slater index and a related parameter - the degree of linearity, $\lambda$, also often used in the literature. It turns out that $\lambda$ may be treated as the mode of the posterior distribution of $p$ if the latter is concentrated in the vicinity of $p=1$ (which would be also an appropriate definition of rankability). In simple terms, and without surprise, sufficiently large $\lambda$ is a good indicator of rankability; our analysis endows it with a probabilistic interpretation.

We show numerical examples using synthetic and real-world data that illustrate our theoretical findings.

As an additional result, thanks to the new formulation of the estimation problem, we can find \emph{all} optimal solutions of the \gls{lop}. When ordering $M$ objects, this is done with the complexity $O(M2^M)$ which, notably, is the same maximum complexity reported in \cite{Remage66} to find only one solution. The algorithms we develop and use are implemented in Python and made available at \github.

Since our work is methodological in nature and aims at providing theoretical tools to analyze the \gls{lop}, it does not immediately address all the practical/implementation issues. Indeed, the complexity of finding the Slater spectrum grows very quickly with $M$ and becomes impractical for $M>30$, while many practical problems deal with size $M>100$ \citep{Anderson19}, \citep{Ceberio15}. 

Although we made some steps to link our framework to the practical setup (\eg by reinterpreting, in probabilistic terms, the degree of linearity, which can be calculated for large problems), the possibility of finding the Slater spectrum for large $M$ (exactly or approximately), or, at least, to calculate some statistics of $p$ remains the main challenge for wider applicability of the proposed methods.

\ifdefined\ARXIV
\input{\CFilesBib/output.bbl.my}
\else
\bibliography{\CFilesBib/references_rank,\CFilesBib/IEEEabrv,\CFilesBib/references_all}
\bibliographystyle{abbrvnat}
\fi

\end{document}